\documentclass[twoside,12pt]{article}
\usepackage{natbib}
\usepackage{hyperref}
\usepackage[english]{babel}
\usepackage{setspace}
\usepackage{tikz}
\usetikzlibrary{automata,topaths}
\usepackage{xcolor}
\usepackage{bbm}
\usepackage{amsthm,amsmath,amsfonts,amssymb,bbm}
\usepackage{algorithm}
\usepackage{algpseudocode}

\newtheorem{theorem}{Theorem}
\newtheorem{lemma}[theorem]{Lemma}
\newtheorem{definition}[theorem]{Definition}
\newtheorem{remark}[theorem]{Remark}
\newtheorem{corollary}[theorem]{Corollary}
\newtheorem{example}[theorem]{Example}
\newtheorem{proposition}[theorem]{Proposition}

\def\R{\mathbb{R}}
\def\N{\mathbb{N}}

\def\Ebb{\mathbb{E}}

\def\Acal{\mathcal{A}}
\def\Bcal{\mathcal{B}}

\def\Lcal{\mathcal{L}}
\def\Mcal{\mathcal{M}}
\def\Ncal{\mathcal{N}}
\def\Pcal{\mathcal{P}}
\def\Scal{\mathcal{S}}
\def\Tcal{\mathcal{T}}
\def\Ucal{\mathcal{U}}
\def\Vcal{\mathcal{V}}
\def\Xcal{\mathcal{X}}

\def\onebb{\mathbbm{1}}

\def\dfrak{\mathfrak{d}}
\def\gfrak{\mathfrak{g}}

\newcommand{\independent}{\mathrel{\perp\!\!\!\perp}}

\def\algoname#{HEMQ}%the "#" after def means that this command NEEDS to be called as \algoname{}
\newcommand{\blind}{1}%0=blind, 1=un-blind
\begin{document}

\def\spacingset#1{\renewcommand{\baselinestretch}%
{#1}\small\normalsize} \spacingset{1}

%%%%%%%%%%%%%%%%%%%%%%%%%%%%%%%%%%%%%%%%%%%%%%%%%%%%%%%%%%%%%%%%%%%%%%%%%%%%%%

\if1\blind
{
 \title{\bf Huber-energy measure quantization}
\author{Gabriel Turinici\footnote{\text{gabriel.turinici@dauphine.fr}, \url{https://turinici.com}} \\
        CEREMADE,       Universit\'e Paris Dauphine - PSL \\
       Place du Marechal de Lattre de Tassigny 75016 PARIS, FRANCE
       }
  \maketitle
} \fi

\if0\blind
{
  \bigskip
  \bigskip
  \bigskip
  \begin{center}
    {\LARGE\bf Huber-energy measure quantization}
\end{center}
  \medskip
} \fi
%
%\bigskip
%
%\title{Huber-energy measure quantization}
%\maketitle
%
\begin{abstract}%   <- trailing '%' for backward compatibility of .sty file
We describe a measure quantization procedure i.e., an algorithm which finds the best approximation of a target  
probability law (and more generally signed finite variation measure)
by a sum of $Q$ Dirac masses ($Q$ being the quantization parameter). 
The procedure is implemented by minimizing the statistical distance between the original measure and its quantized version; the distance is built from a negative definite kernel and, if necessary, can be computed on the fly and feed to a stochastic optimization algorithm (such as SGD, Adam, ...). We investigate theoretically the fundamental questions of existence of the optimal measure quantizer and identify what are the required kernel properties that guarantee suitable behavior. We propose two best linear unbiased (BLUE) estimators for the squared  statistical distance and use them
in an unbiased procedure, called \algoname{}, to find the optimal quantization.
We test \algoname{} on several databases: multi-dimensional Gaussian mixtures, Wiener space cubature, Italian wine cultivars and the MNIST image database. The results indicate that the \algoname{} algorithm is robust and versatile and, for the class of Huber-energy kernels,  matches the expected intuitive behavior.  
\end{abstract}

%\begin{keywords}
%vector quantization, energy norm, clustering, stochastic optimization
%\end{keywords}

\noindent%

%\vfill
%
\newpage
%\spacingset{1.9} % DON'T change the spacing!

\section{Introduction}
\subsection{Motivation}
Working with uncertainty described as finite variation measures  (such as, for instance, a probability law) is of paramount importance in many scientific fields. However, only rarely analytical solutions can be found and the numerical approaches replace exact objects by some discrete versions. We analyze here a discretization dimension not often considered in the literature, namely the measure quantization i.e. the description of a finite total variation measure $\mu$ on a set $\Xcal$ through a sum of Dirac masses $\delta_{\alpha,X}:=\sum_{q=1}^Q \alpha_q \delta_{x_q}$; here $Q$ is a user-specified integer that acts as a discretization parameter, $\alpha_q$ are real numbers (weights) and $x_q \in \Xcal$ are the locations of the Dirac masses. The  weights $\alpha_q$ and $x_q$ are to be chosen in order to ensure that $\delta_{\alpha,X}$ is close to $\mu$ according to a metric $d(\cdot,\cdot)$ which will be defined later.

Based on best linear unbiased (BLUE) estimators, we construct an unbiased procedure, called \algoname{}, that minimizes the distance 
$d(\mu,\delta_{\alpha,X})$ with respect to $\alpha$ and $X$ or only with respect to $x$ when $\alpha$ is fixed, with the uniform distribution $\alpha_q=1/Q$ being a remarkable example. The distance $d(\cdot,\cdot)$ is built from a negative definite kernel $h(\cdot,\cdot)$ and can be computed on the fly and feed to a stochastic optimization algorithm (such as SGD, Adam, ..). This allows to work with low memory requirements even for high values of $Q$; the kernel analytical properties are tailored to be compatible with modern hardware (like GPUs) that feature important speed-ups at the price of low precision floating point computations.

Although similar approaches have been discussed in the literature (see following section) a general answer to fundamental questions such as the existence of a minimizer 
 is still lacking; to address these issues we give in Sections \ref{sec:existence}
and \ref{sec:further_theory} 
several theoretical results to identify the kernel properties that ensure convenient behavior.
We introduce in Section~\ref{sec:blue_estimator} two estimators 
of the squared distance and prove, for the first time, that they are BLUE.

In Section \ref{sec:numerics} the \algoname{} procedure is then tested on several databases: multi-dimensional Gaussian (and Gaussians mixtures), Italian wines dataset and the MNIST image database. Satisfactory results are obtained that illustrate the potential of this method.

\subsection{Relationship with the literature}

The question of describing a measure by a finite number of Dirac masses has already been addressed in specific contexts in the literature. We give below some entry points to these works.

\subsubsection{Vector quantization}

Literature contains many information and procedures for vector quantization of measures, see \cite{book_quantization_measures,optimal_quantization_wasserstein11,pages_optimal_2018}. In vector quantization the goal is to divide data into clusters, each represented by its centroid point. Some applications are K-means and more general clustering algorithms. 

The difference with our approach is twofold: first the distances involved are not the same: in vector quantization the relevant distance is related to the Wasserstein-Kantorovich metric (cf. \cite{book_quantization_measures}[Section 3, page 30 and page 34] and \cite{optimal_quantization_wasserstein11}~) 
while here we have a kernel-based distance. This gives rise to different theoretic questions; in addition the existence of a kernel makes our computation of the distance very different from the case of vector quantization where the concept of Voronoi diagram is central. Note that the Voronoi digram is intrinsically related to {\bf positive} measures, while in our case the measure has only to have bounded variation.
 This is handy when one has already a partial compression and only want to improve it,  
 and of course for general signed measures.

As a last difference, note that the weights of the Voronoi cells are not known {\it a priori} (i.e., are optimized) while in our approach these weights (denoted $\alpha_q$ above)
can be either considered fixed (e.g., uniform) or subject to optimization.

When the cardinal of the compressed distribution is large this procedure looses efficiency because it needs to take into account the full
set of codevectors and cannot only sample a part of them (see nevertheless \cite{guo_accelerating_2020,ann_benchmarks_site,measure_quantization} for alternative approaches); on the contrary, our procedure %is unbiased and thus 
allows, if necessary, to only work with a sub-sample at the time,  reducing the memory requirements from $Q^2$ (full size of the quantized set) to $B^2$ ($B$ is the batch sample).
\subsubsection{Kernel vector quantization and ``neural gas"}
A related approach is the kernelized vector quantization of 
Vilmann et al., see~\cite{villmann_kernelized_2015} that use self-organizing maps techniques to reach a quantization. 
From the technical point of view they require that the kernel be 
differentiable and universal or that it is based on some kind of divergence.
The quantization at its turn is mainly used for clustering. See also \cite{chatalic_nystrom_2022} that use a stochastic quantization (i.e., the points are random variables).

A related approach is the ``neural gas" algorithm, see~\cite{martinetz1991_neural_gas} that also used centroids and adapts them according to a ``neighborhood" rule.

Also in the general area of clustering, the energy kernel (precisely the one we use in this work as an important particular case), has been used in \cite{szekely2005hierarchical} and \cite{li2015k}; they (citation) ``compute the energy distance between clusters and merge clusters with minimum energy distance at each step". Although this is not formalized as a Hilbert space embedding of Borel measures, the approach is relevant.
They also prove a statistical consistency result in \cite[Section 2.2]{szekely2005hierarchical} 
and show that corrections are necessary for finite batch sample (see Section \ref{sec:numerics} below).

\section{Existence of the optimal measure quantifier}
\label{sec:existence}

We answer in this section the fundamental question of the existence of the optimal measure quantifier for general (albeit fixed) weights. We refer to the 
Proposition~\ref{prop:existence_variable_alpha}
%Proposition~\ref{prop:uniqueweights} 
for results on the optimization when the weights are variables too  (see also Example~\ref{ex:alpha_variable} and various remarks below).

\subsection{Notations}

Consider $\Xcal$ to be a domain in the $N$-dimensional space $\R^N$; we will work with measures with support included in $\Xcal$.
The set of all measures (including signed measures) with support in $\Xcal$ having finite total variation
will be denoted $\Tcal\Vcal(\Xcal)$ while the probability laws will be denoted 
$\Pcal(\Xcal)$.
We denote 
\begin{equation}
TV(\zeta) := \int |\zeta|(dx), \forall \zeta \in \Tcal\Vcal(\Xcal).
\end{equation}
We also suppose for convenience that $0\in \Xcal$ but this hypothesis is not required and one can replace $0$ by some arbitrary point in  $\Xcal$ and all results will still hold.
We consider
$k : \Xcal \times \Xcal \to \R$ a (real) symmetric positive kernel and 
 $h$ its associated squared distance function 
\begin{equation}
 h(x,y)= k(x,x)+k(y,y)-2k(x,y).
 \label{eq:relation_h_function_of_k}
\end{equation}
Note that $h(x,x)=0$, $\forall x \in\Xcal$. 
 By convention 
 %when \eqref{eq:relation_h_function_of_k} is true 
 we will also note $h(x) = h(x,0)$. 
\\
Throughout this section, 
let $Q$ be a fixed positive integer. 
Denote 
\begin{equation}
    \Pcal_Q =  \left\{
    (\beta_q)_{q=1}^Q\in\R^Q :  \beta_q \ge0, \ \sum_{q=1}^Q \beta_q =1 \right\},
\label{eq:definition_pcal}
\end{equation}
%and fix $\alpha\in \Pcal_Q$.
For any  $k$, $h$ such that \eqref{eq:relation_h_function_of_k}
holds and $z\in \Xcal$ arbitrary but fixed 
 denote (see Appendix \ref{sec:appendix_embedding}  Lemma~\ref{lemma:properties_kernels} item 4)~:
\begin{align}
&\Mcal^h =	\Mcal_k := \left\{ \mu \in \Tcal\Vcal(\Xcal):
	\int_\Xcal \sqrt{k(x,x)}|\mu|(dx) < \infty\right\}
	\label{eq:definition_space_finite_measures_kernel}
\\ &
= \left\{ \mu \in \Tcal\Vcal(\Xcal):
\int_\Xcal \sqrt{h(x,z)}|\mu|(dx) < \infty\right\},
\end{align}
The kernel $k$ induces a distance $d$ (cf.~\cite[Eq.~(10)]{sriperumbudur2010hilbert}) defined for any two signed measures $\eta_i \in \Mcal_k$, $i=1,2$ by~:
\begin{equation}
	d(\eta_1,\eta_2)^2 = \int_\Xcal \int_\Xcal  k(x,y) (\eta_1-\eta_2)(dx) (\eta_1-\eta_2)(dy).
	\label{eq:defdistance_k}
\end{equation}
Note that invoking \eqref{eq:inequality_kxy_kxxkyy}~:
\begin{align}
& d(\eta_1,\eta_2)^2 \le \int_{\Xcal\times \Xcal} |k(x,y)| |\eta_1-\eta_2|(dx) |\eta_1-\eta_2|(dy) \nonumber \\ 
& \le	 
\int_{\Xcal\times \Xcal} \sqrt{k(x,x)} \sqrt{k(y,y)} (|\eta_1|+|\eta_2|)(dx) (|\eta_1|+|\eta_2|)(dy)
\nonumber \\
&= \left(\int_{\Xcal} \sqrt{k(x,x)} (|\eta_1|+|\eta_2|)(dx)\right)^2 < \infty, 
\label{eq:finite_distanceMk}
\end{align}
the last inequality being true because  $\eta_1,\eta_2 \in \Mcal_k$. 
When 
%If in addition 
%\begin{equation}
%	\int_\Xcal \eta_1(dx) = \int_\Xcal \eta_2(dx)
%	\label{eq:same_mass_eta12} \end{equation}
%and 
$\int_\Xcal k(x,x) |\eta_i|(dx) < \infty$\footnote{This condition is technical see also Lemma \ref{lemma:properties_kernels} Item \ref{item:distance_same_kktilde} where we prove that when 
	$\int_\Xcal \eta_1(dx) = \int_\Xcal \eta_2(dx)$  the distance in \eqref{eq:defdistance_k} only depends on $h$ and not on the specific choice of $k$ that gives this $h$ from \eqref{eq:relation_h_function_of_k}.} for $i=1,2$ 
 from the formula~\eqref{eq:relation_h_function_of_k} we obtain :
\begin{equation}
\text{ If }	
	\int_\Xcal \eta_1(dx) = \int_\Xcal \eta_2(dx) \text{ then: } 
d(\eta_1,\eta_2)^2 = -\frac{1}{2}\int_\Xcal \int_\Xcal  h(x,y) (\eta_1-\eta_2)(dx) (\eta_1-\eta_2)(dy).
	\label{eq:defdistance}
\end{equation}

Because of the mass equality condition in \eqref{eq:defdistance} all our ``quantizations" are chosen to have same total mass as the measure to be quantized.
Since $\left(\int \eta_i(dx)\right) \cdot \delta_0$ is in $\Mcal_k$ for $i=1,2$ 
the condition 
$ \eta_i\in \Mcal^h$ 
implies, by \eqref{eq:finite_distanceMk}, that 
$d\left(\eta_i, \left(\int \eta_i(dx)\right) \cdot \delta_0 \right) < \infty$.
Here $\delta_0$ is the Dirac mass at the origin.
%This results from 	\eqref{eq:defdistance_k} and 	 \eqref{eq:inequality_kxy_kxxkyy}~\footnote{In fact, invoking \eqref{eq:inequality_kxy_kxxkyy}, for any $\mu,\nu \in \Mcal^h$: $d(\mu,\nu)^2 \le \int_{\Xcal\times \Xcal} |k(x,y)| |\mu-\nu|(dx) |\mu-\nu|(dy) \le	 \int_{\Xcal\times \Xcal} \sqrt{k(x,x)} \sqrt{k(y,y)} (|\mu|+|\nu|)(dx) (|\mu|+|\nu|)(dy)= \left(\int_{\Xcal} \sqrt{k(x,x)} (|\mu|+|\nu|)(dx)\right)^2$ which is finite because  $\mu,\nu \in \Mcal^h$.}.
In particular $h(x,y)= d(\delta_x,\delta_y)^2$, $h(x)= d(\delta_x,\delta_0)^2$.
Note that the kernel 
in  \eqref{eq:relation_h_function_of_k} 
induces on  $\Mcal_k$ (see Appendix \ref{sec:appendix_embedding}) a (pre-) Hilbert space structure with 
the property that 
$k(x,y)=\langle \delta_x,\delta_y\rangle$ and such that $d$ in \eqref{eq:defdistance}
is the associated distance $\|\eta_1-\eta_2\|^2= d(\eta_1,\eta_2)^2= 
\langle\eta_1-\eta_2,\eta_1-\eta_2 \rangle$.
\begin{remark}[Convention]
It is important to note that we may only have access to the function $h$ and not to the kernel $k$. Recall \cite{sriperumbudur2011universality} that if $h$ is given many kernels $k$ can be constructed that have the same ``square distance" function $h$ (see also  Appendix~\ref{sec:appendix_embedding}). Accordingly we will prefer, wherever possible, to formulate hypothesis in terms of  the square distance function $h$ (instead of the kernel $k$).

A second argument supports this view: there are examples where the distance has simpler structure than the scalar product, for instance is translation invariant, as one can see from the fundamental example of euclidean spaces.
\end{remark}
Denote 
for any vector $X=(x_q)_{q=1}^Q \in \Xcal^Q$ and $(\beta_q)_{q=1}^Q\in\R^Q$~:
\begin{equation}
	\delta_{\beta,X} := \sum_{q=1}^Q \beta_q\delta_{x_q}.
\end{equation}

\begin{remark}
	As a matter of vocabulary, for $r\in ]0,2]$, $a \ge 0$ we will call 
\begin{equation}
	k^{HE}_{r,a}(x,y):= \frac{(a^2+ \|x\|^2)^{r/2} +(a^2+ \|y\|^2)^{r/2} - (a^2+ \|x-y\|^2)^{r/2} - a^r }{2}
	\label{eq:df_huber_energy_k}
\end{equation}
a (positive) ``Huber-energy" type  kernel, in reference to the ``Huber loss" (see~\cite{huber_robust_1964}) and the pioneering works of Szekely et al. on the  ``energy kernels" (see~\cite{szekely_energy_2013}); the `energy' kernel corresponds to $a=0,r=1$. Through \eqref{eq:relation_h_function_of_k}
its associated squared distance function is~:
\begin{equation}
	h^{HE}_{r,a}(x,y):=(a^2+ \|x-y\|^2)^{r/2} - a^r. 
	\label{eq:df_huber_energy_h}
\end{equation}
Note that $k^{HE}_{r,a}$ and $h^{HE}_{r,a}$
also satisfy \eqref{eq:relation_h_gives_k_z0} for $z_0=0$.
\end{remark}

\subsection{Measure coercivity} \label{sec:measure_coercivity}

To ease the notations and definitions we consider $\Xcal=\R^N$ but note that the results can be extended to $\Xcal\neq \R^N$ at the price of additional details in the treatment of the hypotheses and proofs.
To prove the existence of optimal measure quantization we need some hypotheses on the distance.
\begin{definition}[measure coercivity]
Let $h:\Xcal\times\Xcal\to \R_+$ be a conditionally negative definite function and $d$ the distance it generates through formula \eqref{eq:defdistance}. The function $h$ is called {\bf measure coercive} if, for any integer $J\ge 1$
and any $\beta\in\Pcal_J$ (cf. Definition \eqref{eq:definition_pcal})~:
\begin{equation}
    \lim_{\|X\|\to \infty} d \left( \delta_{\beta,X}, \delta_0 \right)= \infty.
    \label{eq:measure_coercive}
\end{equation}
\label{def:measurecoercive}
In general, a distance $d$ that satisfies 
    \eqref{eq:measure_coercive} is also called {\bf measure coercive}.
\end{definition}
Note that instead of $\delta_0$ any measure at bounded distance from it can be taken. 
For $J=1$ \eqref{eq:measure_coercive} implies $  \lim_{x \to \infty} h(x)= \infty$~; so in particular if $h$ is bounded it is not be measure coercive; see 
Section \ref{sec:existence_gaussian} for theoretical results in this case. %(cf. example \ref{ex:gaussian}).
 On the other hand, the fact that $h(\cdot)$ is unbounded is not sufficient, see Example 
\ref{ex:unbounded_non_coercive}. 

We will see later on that measure coercivity implies the existence of the optimal measure quantization. But checking Definition \ref{def:measurecoercive} is not very easy and we need simpler, sufficient conditions. To this end 
we introduce the following assumption:
\begin{equation}
    \text{There exists } C_L \in \R \text{ such that } \forall x,y\in \Xcal: h(x)+h(y)-h(x,y) \ge C_L.
    \label{eq:hyp_lowerbound_h}
\end{equation}

\begin{remark}
Note that $h$ satisfies \eqref{eq:hyp_lowerbound_h} if and only if it satisfies 
$ d^2(\delta_x,\eta)+d^2(\delta_y,\eta)-d^2(\delta_x,\delta_y) \ge C_L'$ for some $C_L'$ and $\eta \in \Mcal_k$.
\end{remark}

We prove now that the assumption \eqref{eq:hyp_lowerbound_h} is a sufficient condition for measure coercivity:
\begin{lemma}
A function $h$, %modif
associated to a positive definite kernel $k$ by \eqref{eq:relation_h_function_of_k}, 
which satisfies assumption \eqref{eq:hyp_lowerbound_h} and
such that 
$\lim_{x\to \infty} h(x) = \infty$
is measure coercive in the sense of Definition \ref{def:measurecoercive}.
In particular kernels 
%$\|x-y\|^r$ and $(a^2+ \|x-y\|^2)^{r/2} - a^r$
$h^{HE}_{r,a}$
are measure coercive for any $r\le 1$, $a\ge 0$.
\label{lem:lowerbound_implies_coercive}
\end{lemma}
\begin{proof}
See Section~\ref{sec:proof_lemma_lowerbound_implies_coercive}.
\end{proof}
In order to investigate the properties of some particular kernels, we will need the following technical result~:
\begin{lemma}
Let $\dfrak_r$ be the distance corresponding to 
the negative kernel $h_r(x,y)=h^{HE}_{r,0}(x,y)= \|x-y\|^r$, for $0<r<2$ and let $r' \in ]0,r]$. Then there exist two real constants $C^1_{r,r'} >0$ and $C^2_{r,r'} \in \R$ such that, for any  measures $\eta, \mu \in \Tcal\Vcal(\Xcal)$ such that 
$\int (\eta-\mu)(dx) = 0$~:
\begin{equation}
\dfrak_r(\eta,\mu) \ge TV(\eta-\mu)^2 \left\{ 
C^1_{r,r'} \dfrak_{r'}(\eta,\mu) + C^2_{r,r'} \right\}.
\label{eq:estimation_distancesE}
\end{equation}
\label{lemma:dist_r_inequalities}
\end{lemma}
\begin{proof}
See Section~\ref{sec:proof_lemma_dist_r_inequalities}.
\end{proof}
This result allows to prove the coerciveness of distances $\dfrak_r$~:
\begin{lemma}
All distances $\dfrak_r$ are measure coercive for any $0<r<2$.
\label{lemma:dist_r_coercive}
\end{lemma}
\begin{proof}
The coercivity for $r\le 1$ results from the
Lemma~\ref{lem:lowerbound_implies_coercive}
 because all these distances satisfy hypothesis \eqref{eq:hyp_lowerbound_h}.
When $r\ge 1$, from the measure coercivity of  $r'=1$ and Lemma~\ref{lemma:dist_r_inequalities} we obtain~:
\begin{equation}
   \lim_{x\to \infty} \dfrak_r \left( \sum_j \beta_j \delta_{x_j}, \delta_0 \right) \ge 
  \lim_{x\to \infty} C^1_{1,r} \dfrak_1 \left( \sum_j \beta_j \delta_{x_j}, \delta_0 \right) + C^2_{1,r} = \infty,
\end{equation}
hence the measure coercivity of $\dfrak_r$ for $r\ge 1$.
\end{proof}

\begin{remark}{Numerical implementation of $\|x\|^r$}~:
In practice, to find the measure quantization one needs to use optimization algorithms. In general such methods use gradient information and can become unstable for kernels that are not differentiable at the origin; this motivates the use of the more regular Huber-energy kernels $h^{HE}_{r,a}$ (here $a\ge0$ is a constant); other choices are $\frac{\|x\|^2}{\sqrt{a^2+ \|x\|^2}}$. We analyze these kernels in the following.
%positivity of candidates : code
%Wolfram alpha code 
%\url{https://www.wolframalpha.com/input/?i=LaplaceTransform%5B+Exp%5B-t%5D%2Ft%5E%281%2F2%29%2Fsqrt%28pi%29%2C+t%2C+s%5D}
%\verb|LaplaceTransform[ Exp[-t]/t^(1/2)/sqrt(pi), t, s]|
\end{remark}

Denote $\gfrak_a$ the distance induced by the Gaussian kernel $g_\sigma=e^{-\|x-y\|^2/2\sigma^2}$, 
i.e., in order to fix the constants, 
\begin{eqnarray}
& \ & 
\gfrak_\sigma(\delta_x,\delta_y)^2:=1-e^{-\|x-y\|^2/2\sigma^2} = 
\| \delta_x-\delta_y \|^2_{\Mcal^{\gfrak_\sigma}},
\nonumber \\ & \ & 
\langle \delta_x,\delta_y\rangle_{\Mcal^{\gfrak_\sigma}} = g_\sigma(x,y) =e^{-\|x-y\|^2/2\sigma^2}.
\label{eq:gaussian_distance}
\end{eqnarray}

%modif
\begin{lemma}
	For any $\sigma >0$:  $\Mcal^{\gfrak_\sigma} = \Mcal_{g_\sigma}=\Tcal\Vcal(\Xcal)$.
\end{lemma}
\begin{proof}
It is enough to note that the Gaussian kernel is bounded so by definition any $\Tcal\Vcal(\Xcal)$ measure satisfies 	\eqref{eq:definition_space_finite_measures_kernel}.
\end{proof}
\noindent Note that sometimes we will still write   $\Mcal^{\gfrak_\sigma}$ or $ \Mcal_{g_\sigma}$ to recall that we use a specific topology on these spaces and not the canonical $\Tcal\Vcal(\Xcal)$ topology.

\begin{corollary}
The
Huber-energy distance $\dfrak_{r,a}^{HE}$ induced by the
Huber-energy kernel 
$h_{r,a}^{HE}$  is measure coercive for any $a\ge0$, $r\in ]0,2[$.
In addition the following decomposition holds:
 \begin{equation}
 \dfrak_{r,a}^{HE}(\eta_1,\eta_2)^2 =  
  \frac{1}{-\Gamma(-r)}
 \int_0^\infty \gfrak_{1/\sqrt{s}}(\eta_1,\eta_2)^2 \frac{e^{-as}}{s^{r+1}}ds,
\ \forall \eta_i  \in \Mcal_{h_{r,a}^{HE}}, i=1,2.
\label{eq:decompositionHE}
 \end{equation}
\label{cor:huber_energy_distance_coercive}
\end{corollary}
\begin{proof}
See Section~\ref{sec:proof_cor_huber_energy_distance_coercive}.
\end{proof}
\begin{remark}
Similar results hold for 
the reversed Huber-energy distance
 $\dfrak_{r,a}^{RHE}$ induced by the kernel
$h_{r,a}^{RHE}=\left( 
\frac{\|x-y\|^2}{\sqrt{a^2 + \|x-y\|^2}} \right)^{r/2}$.
In this case the writing in the form \eqref{eq:mixture_formula_sqrtHE} below involves the function
$\frac{1}{s} {\mathcal L}^{-1} \left( \frac{d}{dt} \frac{t^{2r}}{(a+t)^r} \right)$ with the operator
 ${\mathcal L}^{-1}$ denoting  the inverse Laplace transform.
\end{remark}

\subsection{Existence of the optimal quantizer for unbounded kernels} \label{sec:theory_existence}

We now turn to the proof of the existence of the optimal quantization. Note that when the kernel is only defined over a compact set $\Xcal$ the existence result follows directly from the continuity of the norm (the lower bound and coercivity are not necessary any more); same argument proves the existence when the weights are also considered variables (compare Propositions~\ref{prop:existence_fixed_alpha} and \ref{prop:existence_variable_alpha}) because the $\Pcal_Q$ is compact. But for general domains,  
the conclusion is not trivial because the optimal points may end up to be at infinity. The fact that $\lim_{x \to\infty} h(x)=\infty$  is not sufficient either. Consider the following example: 
\begin{example} Take $\Xcal=\R$
and let $k^e$ be a unbounded kernel, 
i.e. 
 $\lim_{x \to\infty} k^e(x,x)= \infty$
 and such that
 %$k^e(0,0)=0$ and 
 $k^e(x,x)$ is increasing for $x\ge 0$. Note that in particular this means that  $\lim_{x \to\infty} h^e(x)= \infty$ where $h^e$ is the associated squared distance. Recall that $h^e(0)=0$. Define
\begin{equation}
F_x(\cdot) =
\begin{cases}
k^e(x,\cdot) & \text{ if }  x\le 0 \\
\frac{k^e(x,\cdot)}{1+h^e(x)}+ \frac{h^e(x)}{1+h^e(x)} \left(
\frac{2}{3}\left\{k^e(-2,\cdot)+k^e(-1,\cdot)+k^e(0,\cdot)\right\} - k^e(-x,\cdot) \right) & \text{ if } x \ge 0
\end{cases}.
    \label{eq:counterexample_coercive}
\end{equation}
Denote $K^e:\Xcal \times \Xcal \to \R$ with $K^e(x,y) = \langle F_x,F_y \rangle$. Positivity is immediate and can be proven as in \cite{rkhs_book_berlinet_agnan}[Lemma 1 p.12]; in the RKHS space associated to $K^e$, the 
distance from the measure $\frac{1}{2}( \delta_x+\delta_{-x} )$ to the measure
$\eta=\frac{1}{3}(\delta_{-2} + \delta_{-1} + \delta_0)$ will tend to zero so the optimal $Q=2$ quantization of the measure $\eta$ is 
the limit of  $( \delta_x+\delta_{-x} )/2$ for $x\to \infty$; however there do not exist points $a$ and $b$ such that  $(\delta_a+\delta_b )/2$ is at minimal distance from
$\eta$.%, thus the quantization of $\eta$ with $Q=2$ points does not exist.
\label{ex:unbounded_non_coercive}
\end{example}

% example of kernel that is built on some function set such that limit at infinity equals $K_1/2+K_2/2$. In this set, $K_1/2+K_2/2$ will not have a converged compression with one point because the best compression is at infinity (zero distance!). Also works by modifying to have at infinity $K_{-1}+K_0/2$ where $K_x$ are $L^2$ slowly decaying functions translated by some parameter. 

Thus, additional hypotheses are required in order to have existence of a minimum and not all kernels are equally suitable.
%\end{remark}

\begin{proposition}[Existence~:~fixed positive weights]
Let $Q$ be a fixed, strictly positive integer and 
consider $h$ a negative definite kernel 
constructed from some positive definite kernel $k$ by \eqref{eq:relation_h_function_of_k}
and $\alpha\in (\R_+)^Q$ (fixed). Consider  $\eta \in \Mcal^h$ %\cap \Tcal\Vcal(\Xcal)$ 
%(cf. \eqref{eq:definition_space_finite_measures_kernel}  for the definition of $\Mcal^h$)
%modif
with $\int_\Xcal \eta(dx) = \sum_q \alpha_q$
 and suppose that~:
\begin{enumerate}
%        \item is a bounded variation measure i.e.,
%        $\int |\eta|(dx) < \infty$;
        %with bounded support: $\forall x \in supp(\eta) : \|x\| \le R $
    \item $h(\cdot,\cdot)$ is continuous on $\Xcal \times \Xcal$;
    \item $h$ is measure coercive in the sense of the Definition \ref{def:measurecoercive}.
\end{enumerate}
Then the minimization problem~:
\begin{equation}
 \inf_{X=(x_q)_{q=1}^Q \in \Xcal^Q } d \left(\delta_{\alpha,X} ,\eta \right)^2 
\label{eq:minimization}
\end{equation}
admits at least one solution $X^\star \in \Xcal^Q$.
\label{prop:existence_fixed_alpha}
\end{proposition}
\begin{proof}
See Section~\ref{sec:proof_prop:existence_fixed_alph}.
\end{proof}

\begin{corollary}
The conclusions of Proposition~\ref{prop:existence_fixed_alpha} are true in particular for any kernel
$h_{r,a}^{HE}$ ($a\ge0$, $0 <r<2$) and $\eta \in \Tcal\Vcal(\Xcal)$ such that $|\eta|$ admits an absolute moment of order $r/2$. 
\label{cor:HE_distanceexistence_fixed_alpha}
\end{corollary}
\begin{proof}
This results from 
Corollary~\ref{cor:huber_energy_distance_coercive} and
Proposition~\ref{prop:existence_fixed_alpha} as soon as we prove that  
if $|\eta| \in \Tcal\Vcal(\Xcal)$ admits an absolute moment of order $r/2$ then $\eta \in \Mcal^{h_{r,a}^{HE}}$. The associated positive kernel is $k_{r,a}^{HE}$  so 
to have $\eta \in \Mcal^{h_{r,a}^{HE}}$
we need to prove 
%\begin{equation}
$\int \sqrt{(a^2 + \| x \|^2)^{r/2}- a^r} \cdot |\eta|(dx) < \infty$
%\end{equation}
which is true if $|\eta|$ admits an absolute moment of order $r/2$. 
\end{proof}

We pass now to the proof of the existence for the situation when the 
(positive)
weights $\alpha_q$ can be optimized too. In order to formulate the result we need a technical assumption for a kernel $k$~:
\begin{eqnarray}
& \ & 
\lim_{x\to \infty} k(x,x) = \infty
\label{eq:assumption_norm_kx_infinite}
\\
& \ & \forall y\in \Xcal~:~\exists~b_y <\infty \text{ such that }~|k(x,y)|\le b_y,~\forall x \in \Xcal
\label{eq:assumption_bounded_scalar_product}
\end{eqnarray}
Note that assumptions \eqref{eq:assumption_norm_kx_infinite}
 and \eqref{eq:assumption_bounded_scalar_product} are satisfied by the Huber-energy kernels $h^{HE}_{r,a}$ for all $r\le 1$ and $a\ge0$.

\begin{proposition}[Existence~:~variable positive weights] 
Let $h$ be a negative definite kernel
constructed from some positive definite kernel $k$ by \eqref{eq:relation_h_function_of_k}
 and choose 
$\eta \in \Mcal^h$ such that $\int \eta(dx) >0$.
%with bounded support: $\forall x \in supp(\eta) : \|x\| \le R $
Suppose~: 
\begin{enumerate}
\item $h(\cdot,\cdot)$ is continuous on $\Xcal \times \Xcal$
%\item $h$ is measure coercive in the sense of the Definition \ref{def:measurecoercive};
\item there exists a positive kernel $k$ associated to $h$ (in the sense of 
\eqref{eq:relation_h_function_of_k})
such that 
assumptions \eqref{eq:assumption_norm_kx_infinite}
 and \eqref{eq:assumption_bounded_scalar_product} are satisfied.
%\item $h$ satisfies assumption  \eqref{eq:hyp_lowerbound_h}.
\item there exists a constant $C_L$  such that~: 
\begin{equation}
\forall~x,y\in\Xcal~:~k(x,y)\ge C_L.
\label{eq:lower_bound_kernel_proof_exitstence_variable_weights}
\end{equation}
\end{enumerate}
Then the minimization problem~:
\begin{equation}
 \inf_{X=(x_q)_{q=1}^Q \in \Xcal^Q, \alpha=(\alpha_q)_{q=1}^Q \in (\R_+)^Q, \ \sum \alpha_q = \int \eta(dx) } d \left(\delta_{\alpha,X} ,\eta \right)^2 
\label{eq:minimization_variable_weights}
\end{equation}
admits at least one solution, $\alpha^\star \in (\R_+)^Q$, $X^\star \in \Xcal^Q$.
\label{prop:existence_variable_alpha}
\end{proposition}
\begin{proof}
See Section~\ref{sec:proof_prop:existence_variable_alpha}.
\end{proof}
\begin{remark}
The Proposition~\ref{prop:existence_variable_alpha} applies to the 
Huber-energy kernels $h^{HE}_{a,r}$ for all $r\le 1$ and $a\ge0$.
In fact the conclusion also remains valid for $r\in [1,2[$ but the proof needs some adjustments.
\end{remark}

\begin{example}
When $\alpha$ are not kept fixed but can be optimized, the 
vanishing weights can defy the intuition~: consider $h(x)=\|x\|$; the measure
$d(1/a \delta_{a^2}+ (a-1)/a \delta_0)$ is at distance $1$ from $\delta_0$ even when $a\to \infty$ ! Thus the neighborhood of the origin does not contain only measures with bounded support and this 
prevents the proof of Proposition~\ref{prop:existence_fixed_alpha} 
to work without additional assumptions in this case (there is lack of compactness in the sequence $(X^n)_{n\ge 1}$).
\label{ex:alpha_variable}
\end{example}

%https://www.wolframalpha.com/input?i=plot+abs%28x%2B1%29%5E1.5%2Babs%281-x%29%5E1.5-2*x%5E1.5%2C+for+x+positive

\subsection{Existence of the measure quantization for the Gaussian kernel}
\label{sec:existence_gaussian}

The goal of this subsection (and in fact for the whole section) is to prove that the measure quantization performs according to the intuition. The results are therefore not surprising but what is surprising is the quantity of technical details needed to prove them. %Consider however the following example. 
The following example shows on the other hand that not all intuition is valid and care needs to be present.
\begin{example}
Consider in $\Xcal=\R$ 
the quantization with one point of the  symmetric law 
$\eta= \frac{\delta_{-3/2}+\delta_{3/2}}{2}$ under the Gaussian kernel with parameter $\sigma=1$. 
Simple computations show that the 
optimum point is {\bf not} the Dirac mass at the origin but rather the Dirac mass 
$\delta_{x^\dag}$
at $x^\dag=1.4632$ (a second optimal solution is the Dirac mass at $-1.4632$). Maybe even more surprising is to see that same phenomenon arrives when one tries to quantize with one point the normal law $\eta=\Ncal(0,\tilde{\sigma}^2)$ of parameter $\tilde{\sigma} >1$: the optimum point is not the origin but slightly displaced to the right (second symmetric solution is displaced to the left) depending on $\tilde{\sigma}$.
%https://www.wolframalpha.com/input?key=&i=exp%28-%28x-1%29**2%29%2Bexp%28-%28x%2B1%29**2%29
%
%cf https://www.wolframalpha.com/input?key=&i=exp%28-%28x-1%29**2%29%2Bexp%28-%28x%2B1%29**2%29 for testing code, need to look at the maximum
This does not happen for the Huber-energy kernels which explains, among other reasons, our preference in using them.
\end{example}

First let us remark that, to the best of our knowledge, in the literature there is no general result on the existence of the measure quantization for the Gaussian kernel. We therefore provide it below. Note that the result is not a consequence of the previous assertions because the Gaussian kernel is {\bf not} measure coercive (in the sense of the Definition \ref{def:measurecoercive}) because it is bounded.
The main problem turns out to prove that the infinity cannot harbor optimal quantization points i.e., there is no `escape' to infinity when going towards the minimum.

\begin{proposition}[existence of measure quantization for the Gaussian kernel] 
Consider the Gaussian kernel $\gfrak_a$ defined in~\eqref{eq:gaussian_distance}. Let
$\eta$ be a probability law and fix an integer $Q \ge 1$.
\begin{enumerate}
\item 
For any $\alpha\in \Pcal_Q$ the minimization problem~:
\begin{equation}
 \inf_{X=(x_q)_{q=1}^Q \in \R^{N\times Q} } \gfrak_a \left(\delta_{\alpha,X} ,\eta \right)^2 
\label{eq:min_problem_gaussian_fixed_alpha}
\end{equation}
admits at least one solution $X^\star \in \R^{N\times Q}$.
\item 
The minimization problem~:
\begin{equation}
 \inf_{X=(x_q)_{q=1}^Q \in \R^{N\times Q}, \ \alpha \in \Pcal_Q } \gfrak_a \left(\delta_{\alpha,X} ,\eta \right)^2 
\label{eq:min_problem_gaussian_variable_alpha}
\end{equation}
admits at least one solution $X^\dag \in \R^{N\times Q}$, $\alpha^\dag\in\Pcal_Q$.
\end{enumerate}
\label{prop:existence_gaussian_kernel}
\end{proposition}
\begin{proof}
See Section~\ref{sec:proof_prop:existence_gaussian_kernel}.
\end{proof}

\section{Statistical consistency and the best linear unbiased estimator (BLUE) of the  squared distance} \label{sec:blue_estimator}

Let $h$ be a negative definite kernel. The uniformly weighted quantization with $Q$ points of a measure $\mu$ is expressed as minimizing the distance 
$d^2\left( \frac{1}{Q}\sum_{q=1}^Q \delta_{X_q}, \mu \right)$ among all possible choices 
$X \in \Xcal^Q$. This leads to the general fundamental question of finding 
an unbiased estimator for the distance $d^2 (\nu, \mu)$ 
for  arbitrary measures $\nu$ and $\mu$ in $\Mcal^h$. We identify below the best linear unbiased estimator (BLUE) of the squared distance.
It is natural to search for estimators that use as building blocks the distances 
$d^2(\delta_{Y},\delta_Z)$ where $Y\sim\nu$ and $Z \sim \mu$; see also \cite[Section 2.2]{szekely2005hierarchical} for related considerations.

We introduce in the proposition below a minimal variance linear estimator (BLUE) that uses no hypothesis on the distributions $\nu$ and $\mu$. To the best of our knowledge, no similar results are available in the literature.

\begin{proposition}[BLUE for the distance]
Let $h$ be a negative definite kernel, 
$\nu, \mu \in \Mcal^h$ and $d(\cdot,\cdot)$ the canonical distance induced by the kernel $h$ as in \eqref{eq:defdistance}.  
\begin{enumerate}
	\item \label{item:blue2samples} Consider $Q , J \ge 2$ fixed positive integers and 
	 i.i.d samples 
	$X_1,...,X_Q$ from $\nu$ and 
	$X_{Q+1}, ..., X_{Q+J}$ from $\mu$.  Then~: 
	\begin{equation}
		\widehat{d^2}^{\star} :=  \frac{\sum_{q=1}^{Q} \sum_{j=Q+1}^{Q+J} d^2(\delta_{X_q},\delta_{X_{j}})^2}{ (Q-1) \cdot (J-1)}  
		- \frac{\sum_{q,q'=1}^{Q} d^2(\delta_{X_q},\delta_{X_{q'}})^2}{2 Q (Q-1)} 
		- \frac{\sum_{j,j'=Q+1}^{Q+J} d^2(\delta_{X_j},\delta_{X_{j'}})^2}{2 J (J-1)}  
		\label{eq:optimal_blue_estimator_def}
	\end{equation}
	is the {\bf best linear unbiased estimator (BLUE : i.e., it is unbiased and has minimal variance)  of  $d^2 (\nu, \mu)$} in the class of linear estimators  
	\begin{equation}
\left\{  \left. \widehat{d^2}^w:=\sum_{a,b \le Q+J} w_{ab} d^2(\delta_{X_a},\delta_{X_b})	
\right| w = (w_{a,b})_{a,b\le Q+J} \in \R^{(Q+J) \times (Q+J)}
\right\}.
\end{equation}
	\item \label{item:blue1sample} Consider now $Q\ge 1$, $J\ge 2$ fixed positive integers.
	 For the particular case when $\nu$ is a sum of $Q$ Dirac masses $\nu=\sum_{q=1}^Q p_q \delta_{x_q}$
	with $(p_q)_{q=1}^Q \in \Pcal_Q$ 
	consider the i.i.d. sampling $Z_1,...,Z_J \sim \mu$~; then the estimator~: 
	\begin{equation}
	\widehat{d^2}^\dag :=  
	d^2 \left(\nu, \frac{\sum_{j=1}^J  \delta_{Z_{j}}}{J}  \right)
	- \frac{\sum_{j,j'=1}^{J} d^2(\delta_{Z_j},\delta_{Z_{j'}})^2}{2 J^2 (J-1)}  
	\label{eq:optimal_blue_estimator_def_1sample}
\end{equation}
is BLUE in the class of linear estimators  
\begin{equation}
	\left\{  \left. \widehat{d^2}^{u,v}:=\sum_{j=1}^J u_j d^2(\nu,\delta_{Z_j})	
+ \sum_{j,j'=1}^J v_{j,j'} d^2(\delta_{Z_j},\delta_{Z_{j'}})
	\right| u=(u_j)_{j=1}^J \in \R^J, v = (v_{j,j'})_{j,j'=1}^J \in \R^{J \times J}
	\right\}.
\end{equation}
\end{enumerate}
\label{prop:blue_estimator}
\end{proposition}	
\begin{proof}
See Section~\ref{sec:proof_prop:blue_estimator};

\end{proof}

\section{Further theoretical results} \label{sec:further_theory}

We give in this section several other useful theoretical results.

\subsection{Mean distance, decay rate}
\label{sec:convergence}

\begin{proposition}
Let $\mu\in \Pcal(\Xcal)$, $\alpha\in\Pcal_J$ fixed and $X_1,...,X_J$  i.i.d. samples from $\mu$.
\begin{enumerate}
\item 
Then mean distance squared from $\delta_{\alpha,X}$ to $\mu$ is given by the formula:
\begin{equation}
\Ebb \left[ d\left(\delta_{\alpha,X} ,\mu\right)^2  \right]
 =
 \frac{
 \Ebb_{Y,Y'\sim \mu, Y \independent Y'} [h(Y,Y')]}{2} \cdot \sum_{j=1}^J (\alpha_j)^2 
 \label{eq:decay_general}
\end{equation}
\item 
In particular if $\alpha_j=1/J$, $j=1,...,J$ then
\begin{equation}
\Ebb_{X_j \sim \mu, i.i.d} 
\left[ d\left( \frac{\sum_{j=1}^J \delta_{X_j}}{J},\mu\right)^2  \right]
 = \frac{\Ebb_{Y,Y'\sim \mu, Y \independent Y'} h(Y,Y')}{2 J} 
 \label{eq:decay_uniform}
\end{equation}
\item the uniform distribution $\alpha_j=1/J$ reaches the minimum 
of $\Ebb_{X_j \sim \mu, i.i.d}  \left[ d\left(\delta_{\alpha,X} ,\mu\right)^2  \right]$ 
among all possible distributions $\alpha \in \Pcal_J$.
\end{enumerate}
\label{prop:mean_distance_and_decay}
\end{proposition}
\begin{proof}
See Section~\ref{sec:proof_prop:mean_distance_and_decay}.
\end{proof}

\begin{remark}
Note that among possible forms of $\alpha \in \Pcal_J$, the decay rate varies greatly. For instance 
when $\alpha_1=1$ and the  other are zero the mean distance remains constant. On the other hand when $\alpha_j$ are in a geometric sequence i.e. 
$\alpha_j=c_a a^j$ with $c_a$ such that $\sum_j \alpha_j=1$ 
and $a>1/2$
then
the mean distance {\bf does not} tend to zero: 
the mean distance is (up to a constant) $\sum_{j=1}^J (a^j)^2$ which is larger than 
$\sum_{j=1}^J \frac{1}{2^{2j}} = \frac{1}{4} \frac{1-2^{-2J}}{1-2^{-2}}= 3(1-2^{-2J}) \to 3 \ (J\to \infty)$.
\end{remark}

\subsection{Uniqueness of the weights}

\begin{proposition}[Uniqueness of the optimal weights] Let $h$ be a 
negative kernel and $d$ the distance it induces on the set $\Mcal^h$ as described in Appendix~\ref{sec:appendix_embedding}. 
Suppose $X=(x_q)_{q=1}^Q$ is given; then the 
minimization of~:
\begin{equation} %modif
    \alpha \in  \left\{ \beta=(\beta_q)_{q=1}^Q \in \R^Q; \sum_q \beta_q = \int \mu(dx) \right\} \mapsto 
    d(\delta_{\alpha,X},\mu) \in \R
\end{equation}
admits a unique solution 
$\delta_{\alpha^\star,X}$; in particular when all $x_q$ are distinct, the optimal $\alpha^\star$ is unique.
Same holds when $\alpha_q$ are  searched in any convex ensemble such as $(\R_+)^Q$ or $\Pcal_Q$.
\label{prop:uniqueweights}
\end{proposition} 
\begin{proof}
See Section~\ref{sec:proof_prop:uniqueweights}.
\end{proof}

\subsection{Exact solution in 1D for the `energy' kernel}

\begin{proposition}[optimality of quantiles in dimension $1$ for the `energy' kernel]
Consider the `energy' kernel distance (cf~\cite{szekely_energy_2013} also named Radon-Sobolev $H^1$ in ~\cite{turinici_radonsobolev_2021}) given by 
 $d(\delta_x,\delta_y)= |x-y|$ in dimension $N=1$; let $\mu$ be an absolutely continuous probability measure and  $\alpha \in \Pcal_J$ the uniform weights i.e., $\alpha_j=1/J$, $\forall j \le J$.  
Then the minimization problem 
\eqref{eq:minimization} admits an unique solution 
$X^\star$ which is such that 
$X^\star_j$ is the quantile of order $\frac{j-1/2}{J}$
of the law $\mu$.
\label{prop:1D_quantiles}
\end{proposition}
\begin{proof}
See Section~\ref{sec:proof_prop:1D_quantiles}.
\end{proof}
\begin{remark}
A generalization of this proposition holds without the hypothesis of absolute continuity for $\mu$ and for a general choice of the weights $\alpha_j$, but in this case the definition of the optimum, involving the generalized quantiles, is more technical.
\end{remark}

\begin{figure}[!htb]
    \centering
    \includegraphics{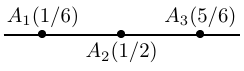}
    \caption{Illustration of the Proposition~\ref{prop:1D_quantiles}.
    The points shown are the quantiles required for the 
    quantization with $Q=3$ points of a 1D law.}
    \label{fig:quantiles_1D}
\end{figure}

We illustrate in Figure~\ref{fig:quantiles_1D} an example of quantiles invoked for the $Q=3$ quantization.

\begin{remark}
In 1D and for the uniform law on the unit segment $[0,1]$ we recover as optimal the points $x_j=\frac{j-1/2}{J}$ which are exactly the optimal points of the vector quantization, cf. \cite{book_quantization_measures}[Section 4.4 page 52]. In particular the decay of the quantization error has similar behavior.
\end{remark}

\subsection{Non convexity of the loss function}

\begin{figure}[!htb]
    \centering
    \includegraphics[width=0.45\textwidth]{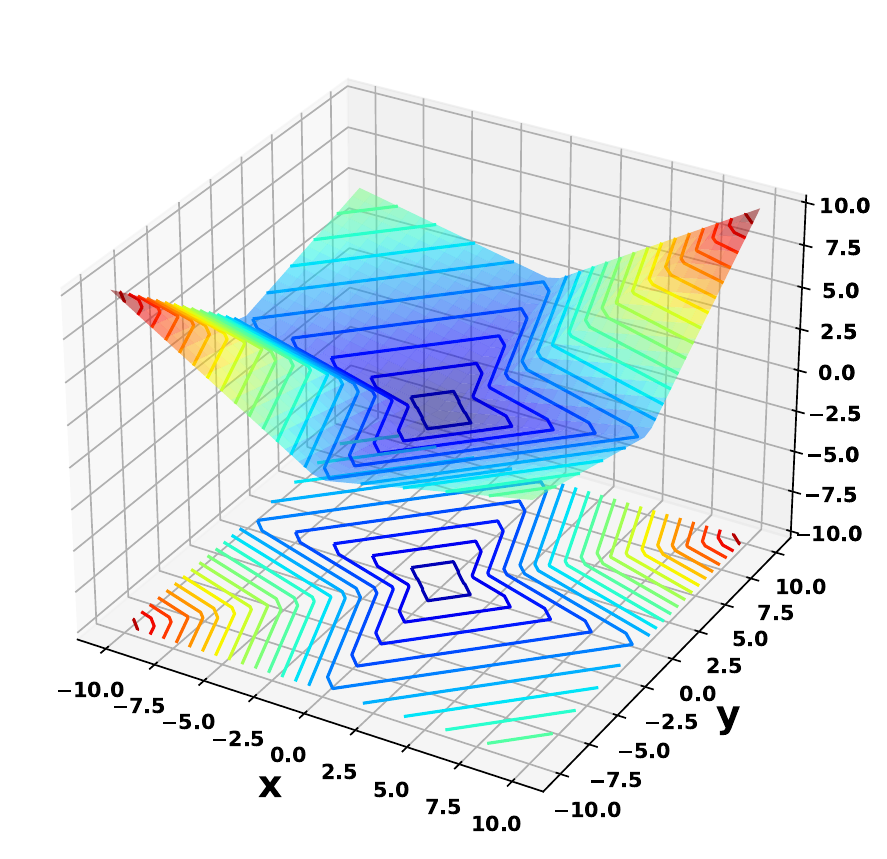}
    \includegraphics[width=0.45\textwidth]{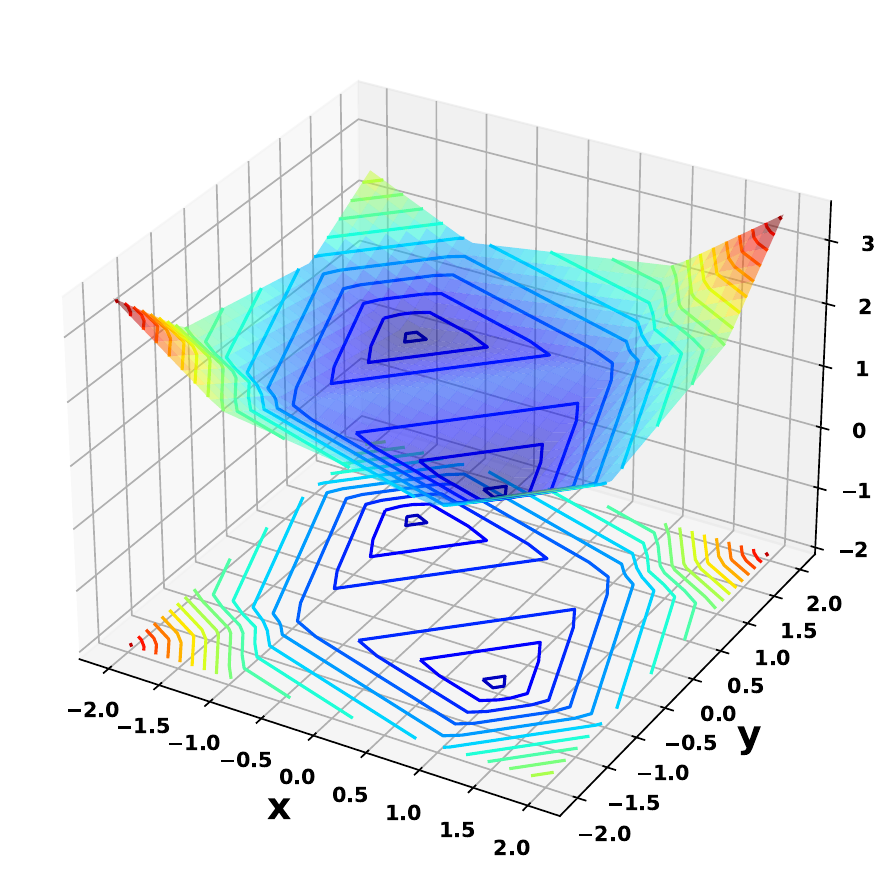}
    \caption{ Graphical representation of $f(x,y) = d\left(\frac{\delta_x+\delta_y}{2},\nu\right)^2$. In both cases $f$ is not a convex function.
    {\bf Left~: for $\nu_0=\delta_0$ we have $f(x,y)=\frac{2|x|+2|y|-|x-y|}{4}$} {\bf Right:} 
    for $\nu_\pm=\frac{\delta_{-1}+\delta_{1}}{2}$ we have $f(x,y)=\frac{2|x+1|+2|x-1|+2|y+1|+2|y-1|-|x-y|-2}{4}$
    }
    \label{fig:nonconvex}
\end{figure}

In order to illustrate better the nature of the optimization problem that we face here, we give below an example that shows that the loss function, taken as the square of the distance from the discrete candidate to the target, is {\bf not} convex with the respect of the location of the quantization points; see also similar remarks from the literature~\cite[page 2]{book_quantization_measures}.
\begin{example}
Consider $N=1$, the energy kernel, the quantization with $Q=2$ points and uniform weights $\alpha=(0.5,0.5)$ and the function
$f(x,y) = d\left(\frac{\delta_x+\delta_y}{2},\nu\right)^2$. We plot in Figure~\ref{fig:nonconvex} the function $f$ for two examples of $\nu$~: $\nu_0=\delta_0$ and 
$\nu_\pm=\frac{\delta_{-1}+\delta_{1}}{2}$.
\end{example}

\section{Numerical results} \label{sec:numerics}
The numerical examples are split into two cases: deterministic optimization and stochastic. Implementations are available in 
the GitHub repository \cite{gabriel_measure_compression_2022}.
\subsection{Deterministic optimization: gradient flow and beyond}
Suppose that we are in a situation when an analytic formula for the mapping $g_\mu : X \mapsto g_\mu(x) = d(\delta_x,\mu)^2$ exists (and also for $ \frac{d}{dx}g_\mu(x)$). Note that such formulas do exist for several situations including the `energy' kernel and $\mu$ a Gaussian  where it reduces to the computation of the first order non-central moment of the Gaussian law 
(see \cite{szekely_energy_2013} and
 \cite[formulas 16, 17, 21 p.299, appendix A.3 p. 303]{turinici_radonsobolev_2021}).
 In this case the optimization of $X$ and / or $\alpha$
 reduces to the optimization of  
 \begin{equation}
  d(\delta_{\alpha,X},\mu)^2 = \sum_{q=1}^Q \alpha_q  g_\mu(X_q)
  - \frac{1}{2} \sum_{q,q'=1}^Q \alpha_q \alpha_{q'} h(X_q,X_{q'}) - \frac{1}{2}\Ebb_{Y,Y'\sim \mu, Y \independent Y'}\|Y-Y'\|.
 \end{equation}
 The last term is a constant and can be neglected.
 The optimization with respect to $\alpha$ is immediate (it is a quadratic problem under the possibly additional constraints $\sum_q \alpha_q = \int_\Xcal\mu(dx)$). The optimization with respect to $X$ can be tackled as a general deterministic optimization procedure.
 
 For some particular cases one can even let converge the following ODE~:
 \begin{equation}
\frac{dX(t)}{dt} = - \frac{d}{dt}     d(\delta_{\alpha,X(t)},\mu)^2,
 \end{equation}
 or even, in order to obtain exponential convergence, use~:
 \begin{equation}
\frac{dX(t)}{dt} = - d(\delta_{\alpha,X(t)},\mu)^2 \frac{\nabla_X [d(\delta_{\alpha,X(t)},\mu)^2]}{\|\nabla_X [d(\delta_{\alpha,X(t)},\mu)^2] \|^2 + \epsilon_{tol}}, \ \epsilon_{tol}=10^{-14}.
\label{eq:exponential_decay_ode_normal}
 \end{equation}
The constant $\epsilon_{tol}$ is introduced to avoid division by zero and depends on the floating point precision of the machine. Note that when 
$\epsilon_{tol}=0$ 
 equation \eqref{eq:exponential_decay_ode_normal} has solution $d(\delta_{\alpha,X(t)},\mu)^2= e^{-t} \cdot d(\delta_{\alpha,X(0)},\mu)^2$,  i.e. exponential convergence is obtained. We give two examples of use below.

\subsubsection{Uniformly weighted low dimension normal quantization}

\begin{figure}[!htb]
\centering
\includegraphics[width=0.45\textwidth]{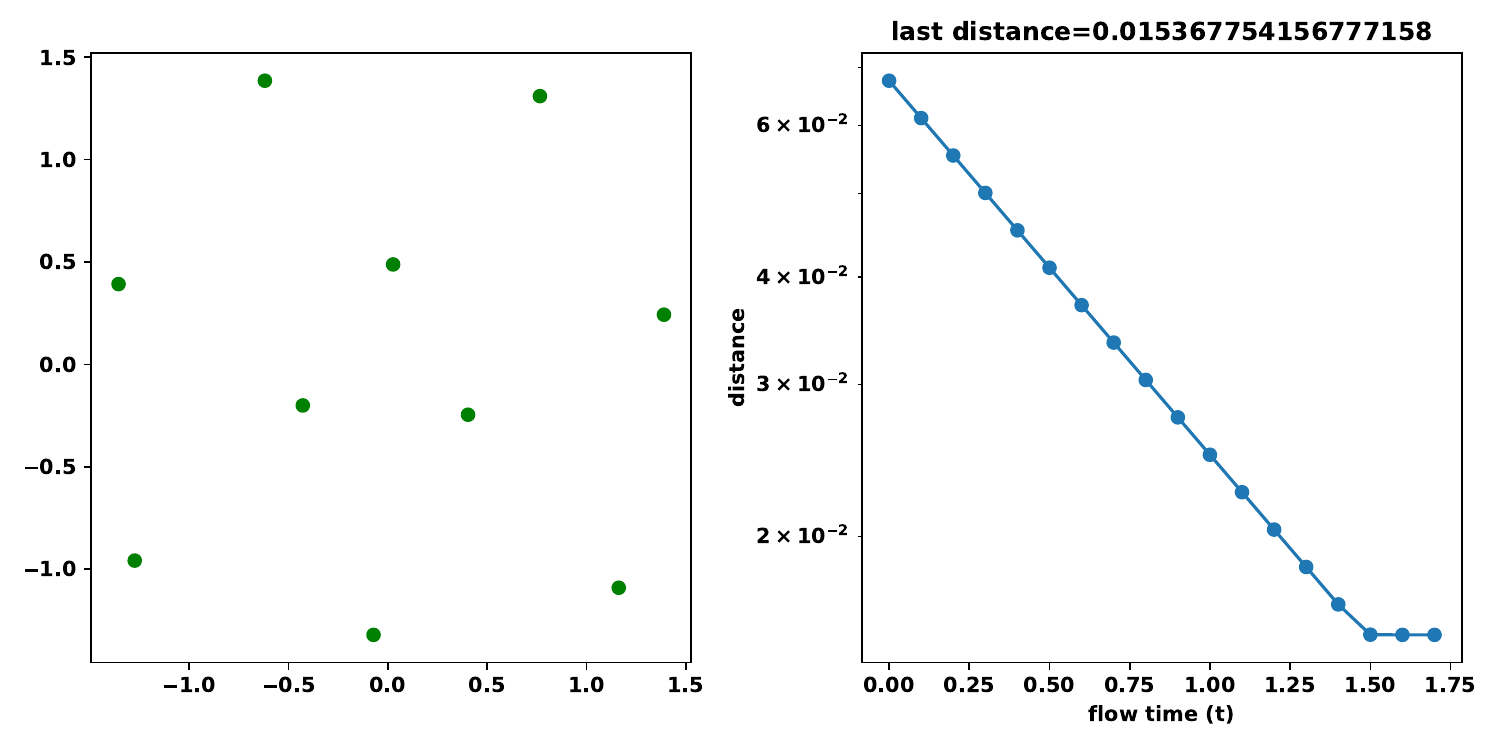}
\includegraphics[width=0.45\textwidth]{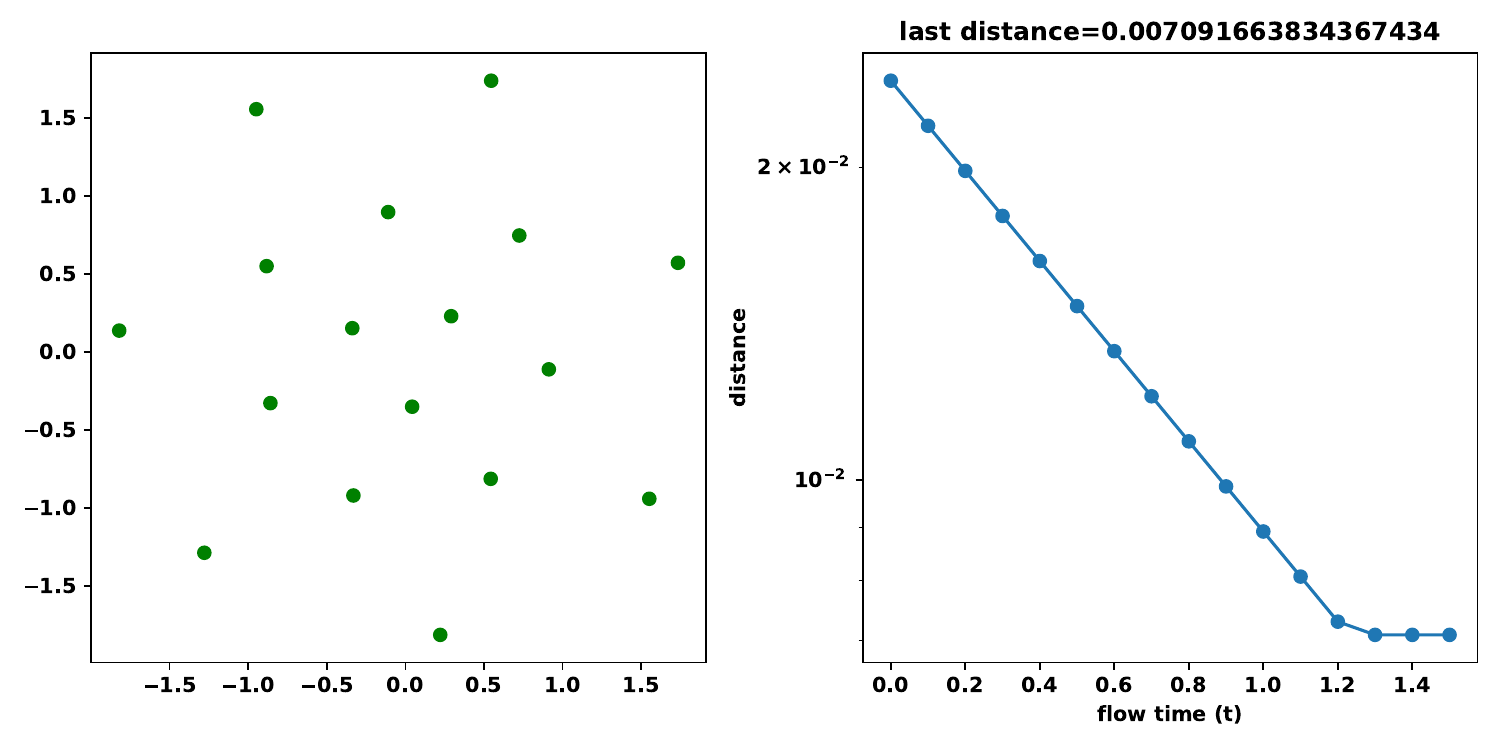}
\includegraphics[width=0.5\textwidth]{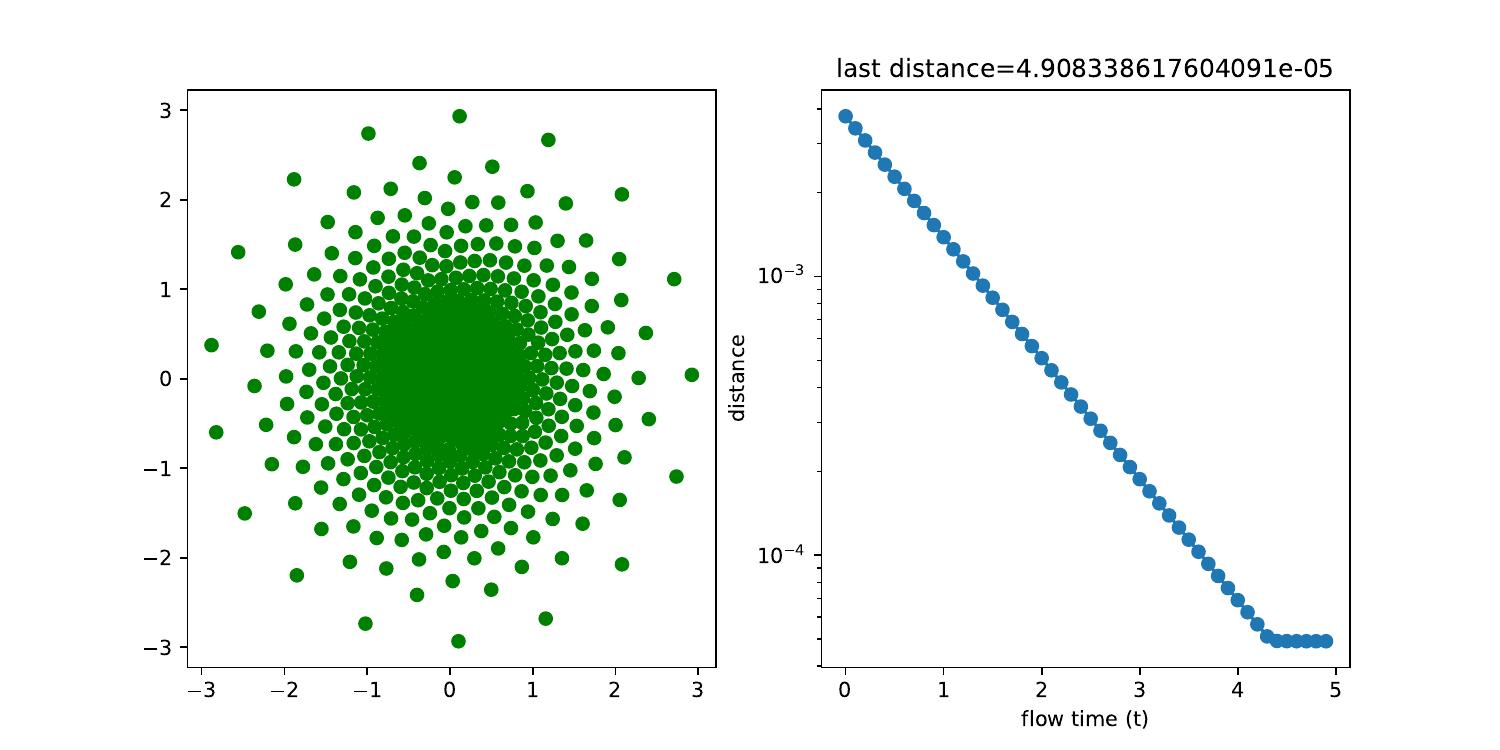}
\includegraphics[width=0.25\textwidth]{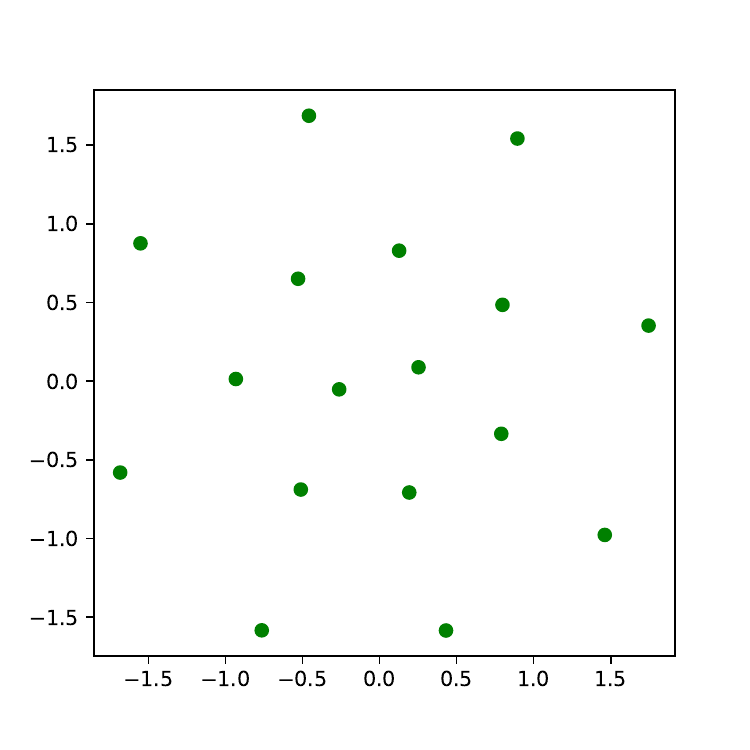}
\caption{Quantization of the bi-variate standard normal distribution ($N=2$) with $Q=10$ (left upper panel) $Q=17$ (upper and lower right panels) 
and $Q=500$ (lower left panel)
points using the evolution in \eqref{eq:exponential_decay_ode_normal}. As expected from the theoretical insights, exponential decay of the distance is obtained.
Total simulation time was set to $T=1.75$ for $Q=10$ and $T=1.5$ for $Q=17$. 
In all cases interesting natural structures appear automatically: when the normal is quantized with $Q=10$ points we observe two concentric rings, one consisting of $3$ points and the other of $7$ points. When $Q=17$ three such rings appear of $2$, $7$ and $8$ points respectively. In general these structures may not be unique as illustrated in the bottom panel where the decomposition is different.
We also plot the convergence of the distance squared.}
\label{fig:2D_gaussian_ode_quantization}
\end{figure}

We take the target to be a multi-dimensional Gaussian. In order to be able to visualize the result we use a 2D Gaussian ($N=2$).
The complete implementation together with an animated illustration are available 
in \if1\blind{the GitHub repository \cite{gabriel_measure_compression_2022}}\fi
\if0\blind{the attached submission files (to be replaced by a github repository)}\fi
 and the results and choices of parameters are given in Figure~\ref{fig:2D_gaussian_ode_quantization}. The result conforms to the intuition and is obtained automatically.

\subsubsection{High dimension normal uniformly weighted quantization as Wiener space cubature}

\begin{figure}[!htb]
\centering
\includegraphics[width=0.3\textwidth]{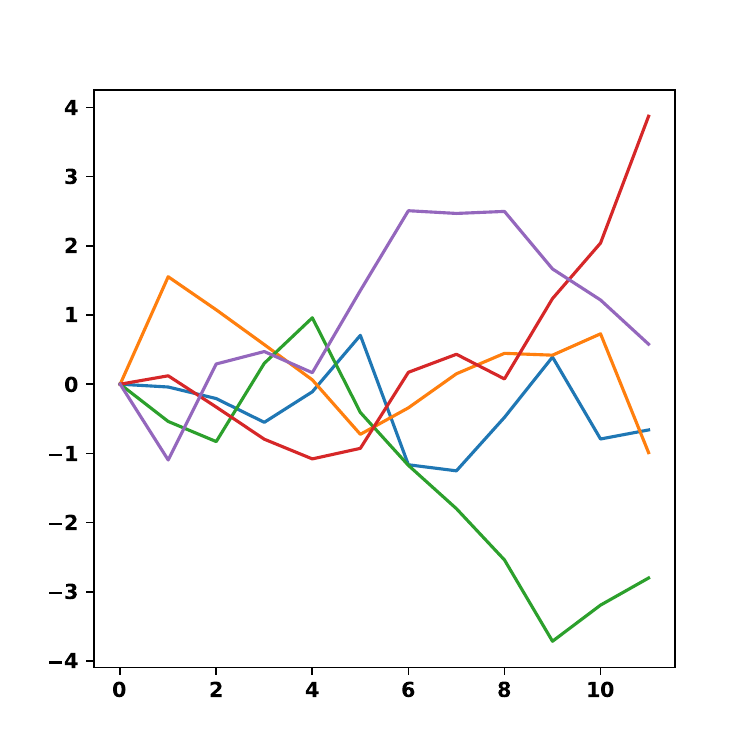}
\includegraphics[width=0.3\textwidth]{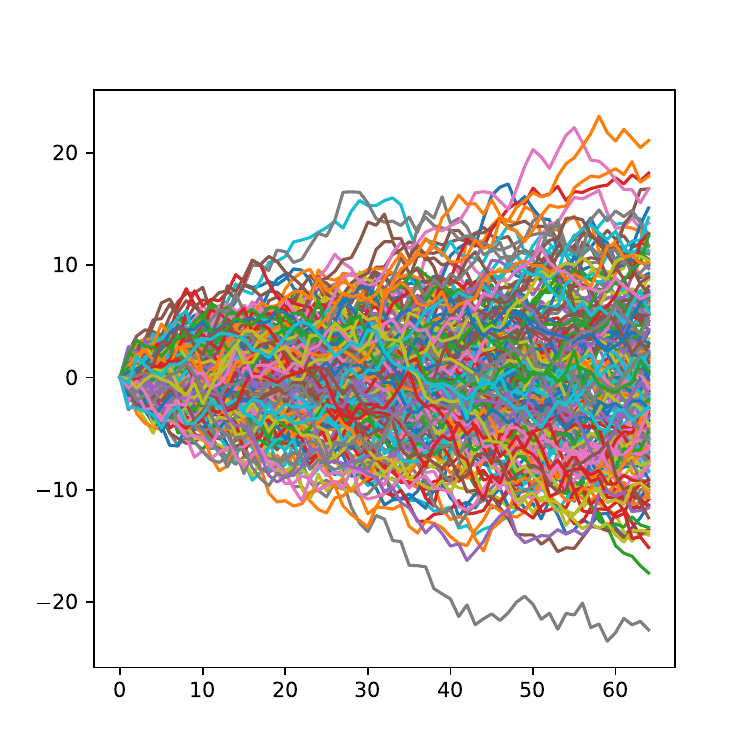}
\includegraphics[width=0.3\textwidth]{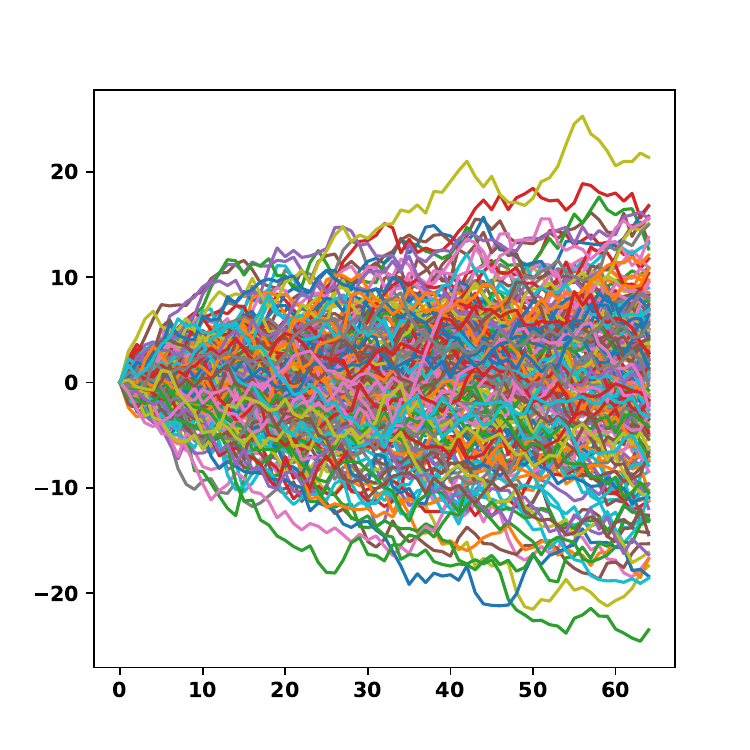}
\caption{Quantization of high dimensional normal distribution. Left panel: $N=11, Q=5$, center panel $N=64, Q=200$. In the right panel we plot a true Brownian simulation (sampling a $N$ dimensional Gaussian). See
 \if1\blind{the GitHub repository \cite{gabriel_measure_compression_2022} }\fi
 \if0\blind{the attached submission files (to be replaced by a github repository) }\fi
 for the implementation.} 

\label{fig:cubature_example}
\end{figure}

Another interesting example is the quantization of a high dimensional normal variable (the covariance being the identity matrix). In order to illustrate graphically the results we employ a trick used often in quantization~: instead of showing a sample of points drawn from a $N$ dimensional Gaussian, we show the Brownian trajectory having the individual coordinates as (independent) increments; quantization in this case is a particular case of cubature on the Wiener space, see \cite{lyons_cubature_2004,pages_optimal_2018} for useful references. An illustration of the results is given in Figure~\ref{fig:cubature_example}. Higher the dimension more difficult is to distinguish the quantization from a real Brownian.

\subsection{Stochastic optimization, part I: a variable weight example}

\begin{figure}[!htb]
\centering
\includegraphics[width=0.5\textwidth]{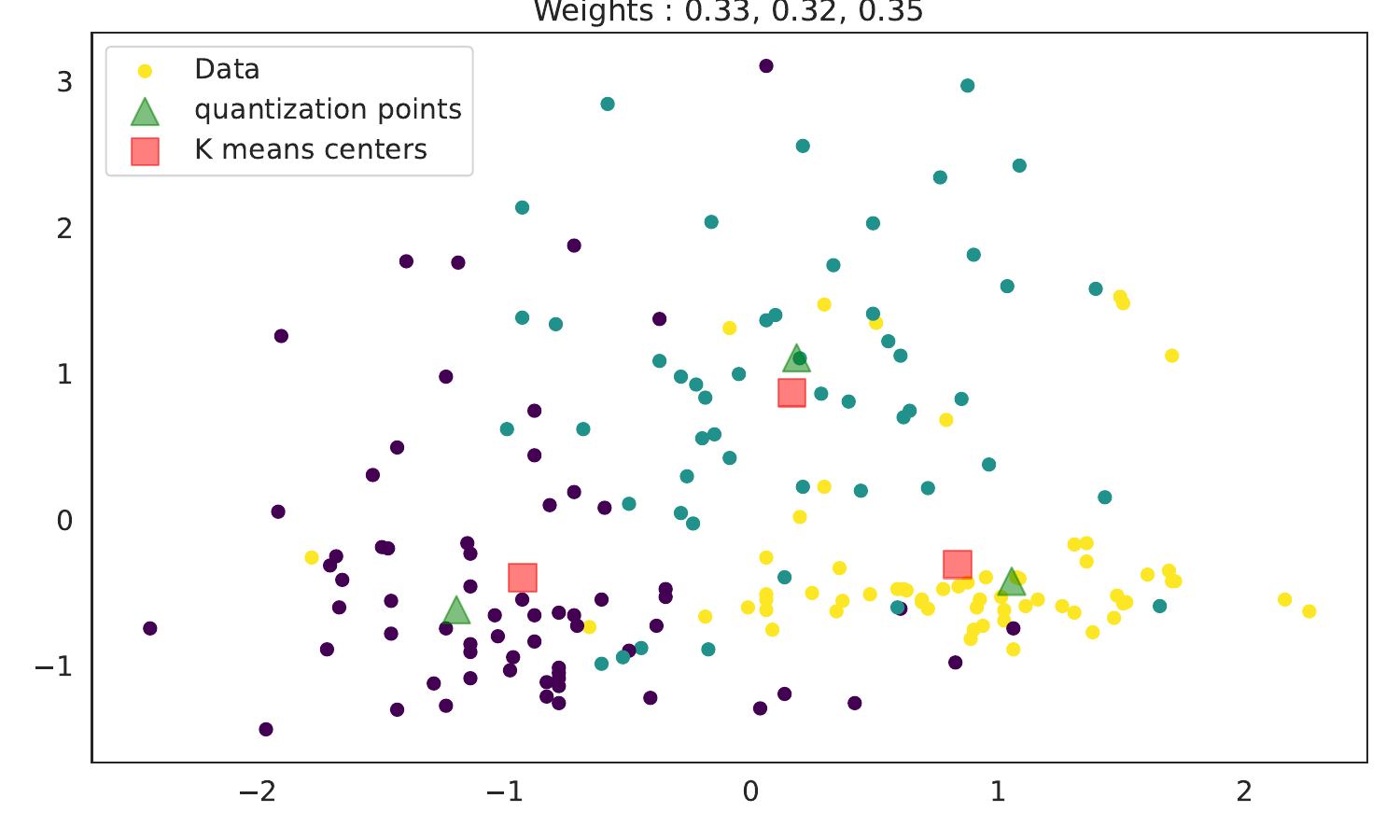}
\caption{Quantization for the ``Italian wines" benchmark~\cite{wines_benchmark}
  using the `energy' kernel.
 Each data point has $13$ dimensions. On each dimension a standardization was performed. We plot a projection on the first two dimensions. The original data points are in solid circles (colored according to their attributed class), the K-means points are in solid squares and the $Q=3$ quantization points are in triangles. The $\alpha$ parameters are given in the title; note that $\alpha$ is {\bf not} supposed to correspond to the class distribution. Python implementation is available in
     \if1\blind{the GitHub repository \cite{gabriel_measure_compression_2022} }\fi
 \if0\blind{the attached submission files (to be replaced by a Github repository) }\fi.}
    \label{fig:italian_wines}
\end{figure}

In order to test the behavior of the (uniformly weighted) quantization with respect to classical machine learning tasks, we compared it with K-means clustering on the UCI repository Italian Wines benchmark~\cite{wines_benchmark}. 
The label data was not used in training (but only for the comparisons). The quantization algorithm was asked 
to optimize both the weights $\alpha_1,\alpha_2,\alpha_3$ and the three points $X_1,X_2,X_2 \in \R^{13}$ to solve the minimization problem \eqref{eq:minimization_variable_weights} for the `energy' kernel.
We employ a non-deterministic algorithm called ``differential evolution" \cite{storn_differential_evolution_1997} (as implemented in Scipy~\cite{virtanen_scipy_2020} version 1.9.1) which has the advantage to not require the gradient, only the distance.  
Once the procedure converged, we took for each points in the dataset the closest quantization point and attributed a class label. As it turns out, our results match {\bf exactly} the class attribution of the K-means algorithm (see ~\cite[Section 7.1]{li2015k}; we recall in table~\ref{tab:confusion} the confusion matrix of the K-means algorithm and display the confusion matrix between the quantization and the K-means.
The results obtained are illustrated in Figure~\ref{fig:italian_wines}. On the other hand note that if we use the Gaussian kernel ($\sigma=1$) the results (not shown here but available on the Github repository) do not coincide any more with those of the K-means procedure. 

\begin{table}[!htb]
    \centering
    \begin{tabular}{|c|ccc|c|   }
    \hline    Class & 1&2&3& Cases \\
    \hline \hline Cultivar I & 59&3&0&62\\
    Cultivar II & 0&65&0&65\\
    Cultivar III & 0&3&48&51\\
    \hline Total & 59&71&48&178 \\ \hline
    \end{tabular}
    \begin{tabular}{|c|ccc|c|   }
    \hline    Class & 1&2&3& Cases \\
    \hline \hline 1 & 62&0&0&62\\
    2 & 0&65&0&65\\
    3 & 0&0&51&51\\
    \hline Total & 62&65&51&178 \\ \hline
    \end{tabular}
    \caption{Classification of Wine Data by K-means : {\bf left~:} confusion matrix of the K-means algorithm, table taken from \cite{li2015k} and reconfirmed by our computations.
     {\bf right~:} confusion matrix between the K-means and the measure quantization algorithm   using the `energy' kernel.
    The classes were relabeled to match the original label names in the data.}
    \label{tab:confusion}
\end{table}
%and the Rand index (cf wiki )
%https://en.wikipedia.org/wiki/Rand_index
%https://link.springer.com/referenceworkentry/10.1007/978-3-0348-0667-1_40
%https://www.kaggle.com/datasets/harrywang/wine-dataset-for-clustering
%https://archive.ics.uci.edu/ml/datasets/wine
%many already done  cf. (see commented latex below)
%https://www.kaggle.com/code/biubiulin/wine-dataset-for-clusteringh
%https://www.kaggle.com/datasets/harrywang/wine-dataset-for-clustering/code

\subsection{Stochastic optimization, part II} \label{sec:stochastic_algo}

In order to go beyond distributions that can be treated semi-analytically we suppose here that we can only sample from the target measure $\mu$ to be quantized. Therefore the optimization is intrinsically stochastic, with the distance being computed on the fly. 
We use the Adam algorithm, see~\cite{kingma_adam_2017}
and employ a learning rate of $0.1$ and (with the notations in the reference) $\beta_1=0.9$, $\beta_2=0.999$.
The procedure \ref{alg:tcalgo} was implemented as indicated below.

\begin{algorithm}
	\caption{Stochastic Huber-energy measure quantization algorithm}
	\label{alg:tcalgo}
	\begin{algorithmic}[1]
		\Procedure{S-HEMQ}{}%{batch size $K$, latent dimension $N$}
		\State $\bullet$ set batch size $B$, parameters $a$ (default $10^{-6}$), and $r$ (default $1.$)
		\State $\bullet$ initialize points $X=(X_q)_{q=1}^Q$ sampled i.i.d from $\mu$
		\While{(max iteration not reached)}
		\State $\bullet$ sample $z_1,...,z_B \sim \mu$ (i.i.d).
		\State $\bullet$ compute the loss		
		$L(X) := \dfrak_{r,a}^{HE}\left( \frac{1}{Q}\sum_{q=1}^Q \delta_{X_q}, 
		\frac{1}{B}\sum_{b=1}^B \delta_{z_b} \right)^2$;
		\State $\bullet$ backpropagate the loss $L(X)$ and use a stochastic algorithm to minimize $L(X)$ and update $X$.	
		\EndWhile\label{euclidendwhile}
		\EndProcedure
	\end{algorithmic}
\end{algorithm}

\begin{remark}[Absence of bias]
Finding the uniformly weighted optimal quantization means  
minimizing the distance 
$\Lcal(X) := \gfrak_{1,a}\left( \frac{1}{Q}\sum_{q=1}^Q \delta_{X_q}, \mu \right)^2$ with respect to $X$; but, cf. Proposition~\ref{prop:blue_estimator} item~\ref{item:blue1sample},
up to terms independent of $X$, $L(X)$ is 
 an unbiased estimate of $\Lcal(X)$ so the algorithm is unbiased.
\end{remark}
\begin{remark}[Memory requirements]
As is, the computation of the loss has memory requirements $(Q+B)^2 \times N$.  
If is too large, estimator \eqref{eq:optimal_blue_estimator_def} provides an unbiased loss 
with memory requirements as low as $(2+2)^2 N$.
\end{remark}

\subsubsection{A mixture example}
\begin{figure}[!htb]
\centering
\includegraphics[width=0.35\textwidth]{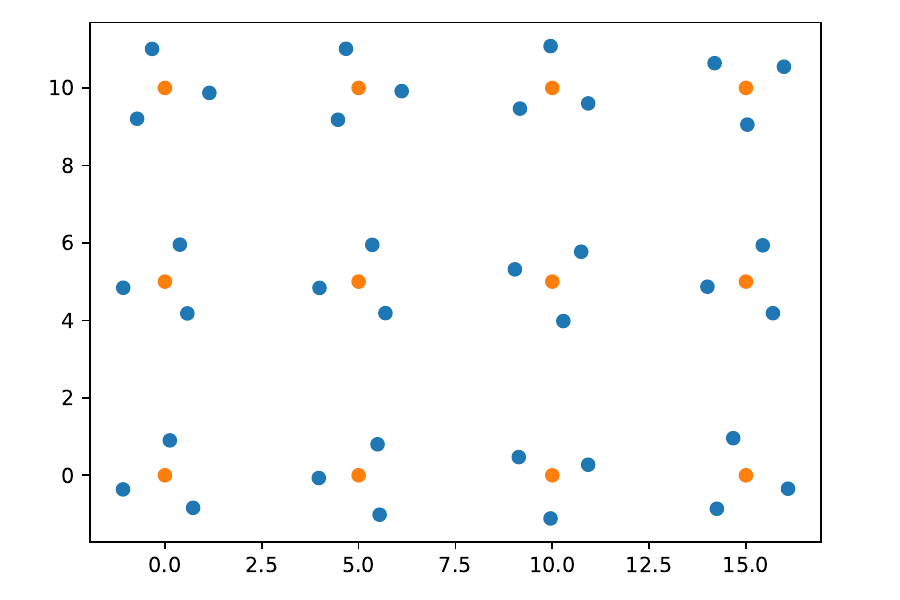}
\caption{Quantization of a mixture of $12$ 2-dimensional Gaussians centered in a $3\times 4$ grid. The number of points is $Q=36$. The result is coherent with the intuition.
}
\label{fig:12g_mixture_example}
\end{figure}

We tested the algorithm \ref{alg:tcalgo} on the target distribution $\mu$ being a mixture of 2D Gaussians
with centers on a $3\times 4$ grid as illustrated in Figure~\ref{fig:12g_mixture_example}.
This measure is quantized with $Q=36$ points.
The algorithm performs well and outputs a result according to intuition by distributing $3$ points to each center; see 
 \if1\blind{the GitHub repository \cite{gabriel_measure_compression_2022} }\fi
 \if0\blind{the attached submission files (to be replaced by a github repository) }\fi
for the implementation. 

\subsubsection{MNIST database sumarizing through quantization}

\begin{figure*}[!htb]
\begin{center}
\includegraphics[width=0.49\linewidth]{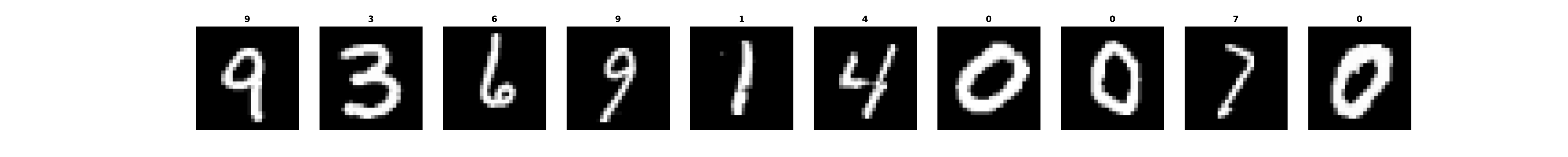}
\includegraphics[width=0.49\linewidth]{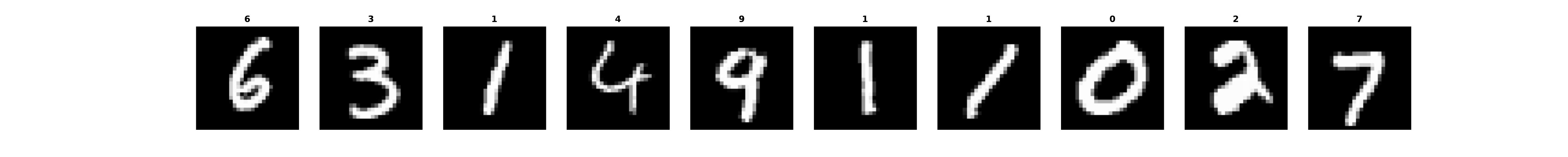}

\includegraphics[width=0.49\linewidth]{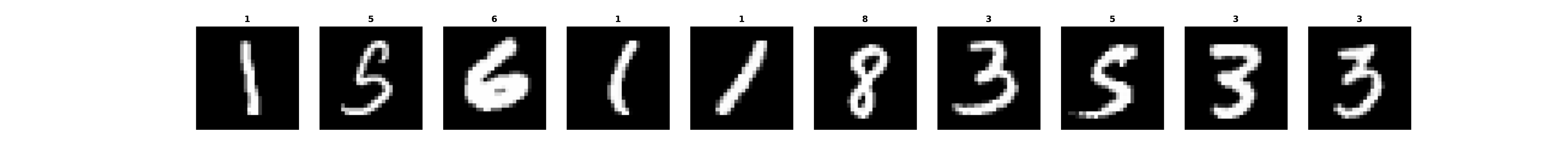}
\includegraphics[width=0.49\linewidth]{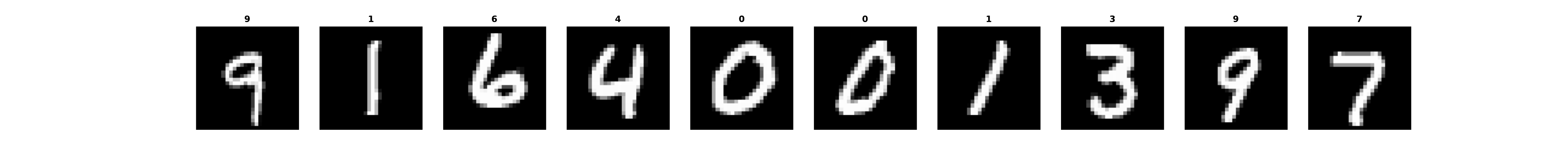}

\includegraphics[width=0.49\linewidth]{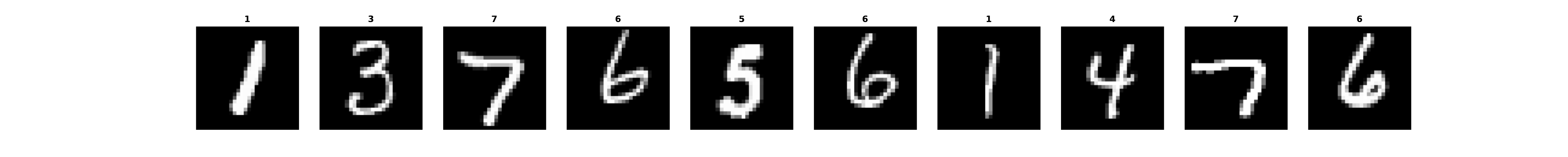}
\includegraphics[width=0.49\linewidth]{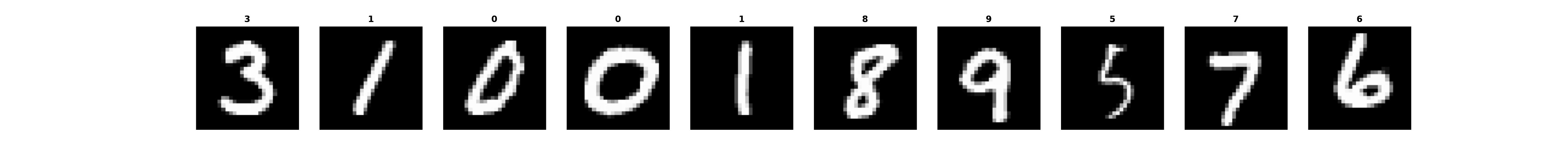}

\includegraphics[width=0.49\linewidth]{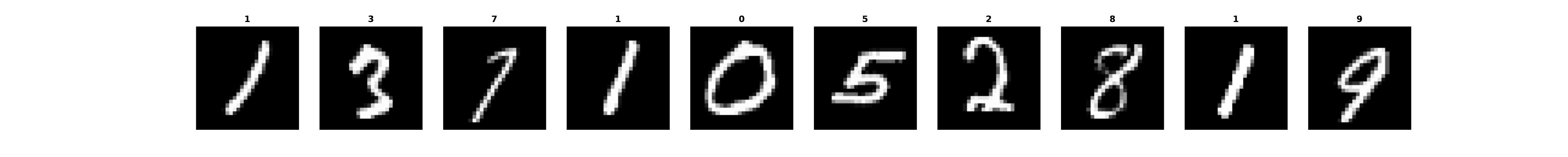}
\includegraphics[width=0.49\linewidth]{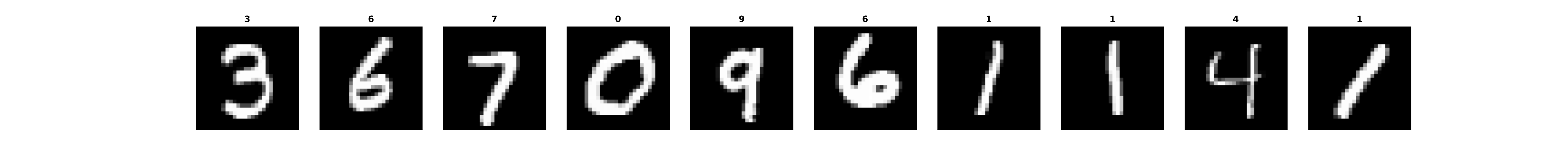}

\includegraphics[width=0.49\linewidth]{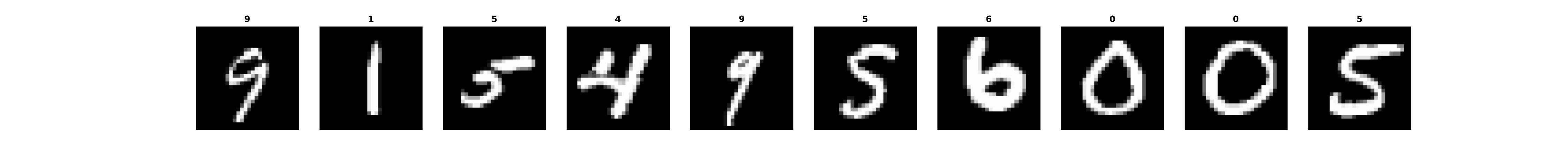}
\includegraphics[width=0.49\linewidth]{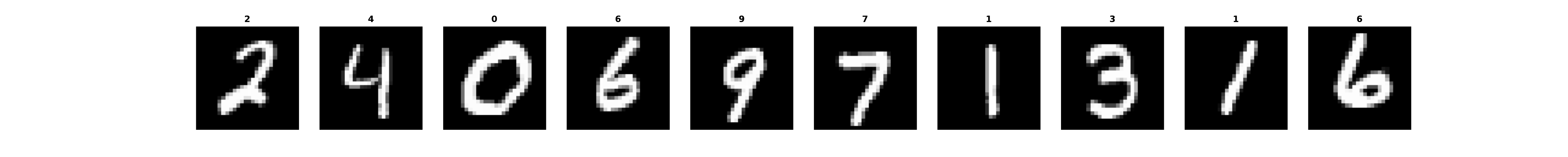}

The DVE metric~:  
\begin{tabular}{c|ccccc|c|c}
row & 1 & 2 &3 &4 &5&mean&std\\\hline
random & 7 & 5 &6&8&6&6.4&1.02\\
quantized & 8 & 7&8&7&8& 7.6&0.49 \\
\end{tabular}

\caption{Five runs of the $Q=10$  MNIST quantization algorithm~\ref{alg:tcalgo} ($a=10^{-5}$, $r=1/2$, $B=100$).
{\bf Left column pictures~:} Independent sampling  from the database; figure repetition is a common feature of these samplings; although statistically likely, high number of repetitions reduces the diversity in the sample.
{\bf Right column pictures~:} Measure quantization mediated sampling. The pictures appear more diverse, for instance have  less repetitions.
{\bf Bottom table~:}
The `Distinct Value Estimation' metric; the results are consistently higher for the quantized samples.
See 
 \if1\blind{the GitHub repository \cite{gabriel_measure_compression_2022} }\fi
 \if0\blind{the attached submission files (to be replaced by a github repository) }\fi
for the implementation. 
}  \label{fig:comparison}
\end{center}
\end{figure*}

MNIST
%~\cite{mnist} 
is a database of 70'000 grayscale $28\times28$ images of handwritten figures. 
%The Fashion-MNIST \cite{xiao2017/online} is a dataset alternative to MNIST consisting of 60k $28\times28$ grayscale images labeled into of $10$ categories: 'T-shirt/top', 'Trouser', 'Pullover', 'Dress', 'Coat', 'Sandal' 'Shirt', 'Sneaker', 'Bag' and 'Ankle boot'; it also has a test set of 10k images.
We used the quantization algorithm~\ref{alg:tcalgo} to extract $10$ ``representative"  images from the database; these points are then projected on the database (i.e., we find the closest one in the database) then compared  with random i.i.d. sampling. The results are presented in \ref{fig:comparison} where we see that the quantization seems to better avoid the repetitions and enforce a more diverse sampling of the database.
Numerical conclusions given in the figure show that the 
``Distinct Value Estimation" (DVE) metric (i.e.
the number of unique figures -- cf.~\cite{haas_sampling-based_1995} and related literature)
is consistently better than the random sampling from the database.
A unilateral t-test confirms (p-value $0.005$) that the DVE mean of optimized sampling is greater than the average DVE value $6.5$ for random sampling (of size $Q=10$) from the MNIST database.

Of course, for the MNIST example we have labels available (that are not used by the quantization procedure) to check a posteriori if a sampling is diverse enough;  but in general the labels are not available and therefore one cannot say whether a given sampling is ``representative" of the distribution or not and has no means to improve it.

\section{Conclusions}

We presented in this work a kernel-based procedure to represent a signed (finite total variation) measure as a quantized sum of (weighted) Dirac masses. We prove some important properties such as the existence of a minimizer and this leads us to consider the Huber-energy class of kernels. Theoretical insights have also been proposed for more ``classical" kernels such as the Gaussian ones. 
The distance is easy to compute and implement, we introduce a BLUE estimator of the squared distance and prove its properties. This leads to propose a quantization procedure (\algoname{}) which is tested with good results on several benchmarks
including multi-D Gaussians, Brownian cubature, Italian wine classification and the MNIST database.
\bibliographystyle{chicago}
\bibliography{measure_quantization_refs}
%\printbibliography

\newpage
\appendix

\section{Appendix: RKHS Kernels, Metric and Hilbert space embedding of measures}
\label{sec:appendix_embedding}
We recall in this section the main concepts and results concerning the reproducing kernel Hilbert spaces (abbreviated RKHS) and how these can help construct metric and Hilbert space structures on ensembles of distributions. We refer the reader to classical books and references for details \cite{schoenberg1938metric,
aronszajn50,
micchelli_interpolation_1986,rkhs_book_berlinet_agnan,sriperumbudur2010hilbert,distance_based_ann_stat2013}.

We recall first the definition of a positive and of a conditionally negative definite kernel on a domain $\Xcal$. Just to be complete, the vocable `kernel' only means `bivariate function' in this context.

\begin{definition}
A symmetric function $\mathbbm{k}:\Xcal \times \Xcal \to \R$ is called a {\bf positive kernel} if:
\begin{equation}
\forall J \in \N \setminus \{0\}, \ \forall \alpha=(\alpha_j)_{j=1}^J \in \R^J, \forall  X=(X_j)_{j=1}^J \in \Xcal^J:
\sum_{j_1,j_2} \alpha_{j_1} \alpha_{j_2} \mathbbm{k}(X_{j_1},X_{j_2}) \ge 0.
\label{eq:definition_positive_kernel}
\end{equation}
The kernel is called {\bf strictly positive definite} if the equality in \eqref{eq:definition_positive_kernel} can only happen when all $\alpha_k$ are zero.
\label{def:positivekernel}
\end{definition}

Note that any positive kernel satisfies~:
\begin{equation}
	\forall x,y \in \Xcal~:~
 |\mathbbm{k}(x,y)| \le \sqrt{ \mathbbm{k}(x,x) \cdot \mathbbm{k}(y,y)}.
 \label{eq:inequality_kxy_kxxkyy}
\end{equation}
This can be proved by checking the Definition~\ref{def:positivekernel}  for points $x,y$ and weights $\alpha,1-\alpha$; we obtain a second order polynomial in $\alpha$ that is always positive. The discriminant condition gives 
 $\mathbbm{k}(x,y)^2 \le \mathbbm{k}(x,x) \mathbbm{k}(y,y)$ hence
 \eqref{eq:inequality_kxy_kxxkyy}.
 
\begin{definition}
A symmetric function $ \mathbbm{h}:\Xcal \times \Xcal \to \R$ is called a {\bf negative kernel}
(also called a {\bf conditionally negative definite kernel})
if:
\begin{eqnarray} & \ & 
\forall J \in \N \setminus \{0\}, \ \forall \alpha=(\alpha_j)_{j=1}^J \in \R^J, \text{ such that } \sum_{j=1}^J \alpha_j = 0,
 \nonumber \\ & \ & \forall  X=(X_j)_{j=1}^J \in \Xcal^J:
\sum_{j_1,j_2} \alpha_{j_1} \alpha_{j_2} \mathbbm{h}(X_{j_1},X_{j_2}) \le 0.
\end{eqnarray}
\label{def:negativekernel}
\end{definition}

\begin{example} In Hilbert spaces 
paramount examples of positive definite kernels are scalar products
$(x,y) \mapsto \langle x, y \rangle$ while 
the distances squared $(x,y) \mapsto \|x -y \|^2$ are  remarkable examples of negative definite kernels.
\end{example}

%v4
%Another source of examples of negative definite kernels is the formula \eqref{eq:relation_h_function_of_k} : if $k$ is a positive  kernel then $h$ given by \eqref{eq:relation_h_function_of_k}  is a negative definite kernel, see \cite[Property 21.5.4 p. 529]{book_dist_probas13}; moreover this kernel satisfies $h(x,x)=0$ for all $x\in \Xcal$.
%
%
Some ways to construct positive and negative kernels and some information on their boundedness are given in the following~:
\begin{lemma}
	Let $\mathbbm{k}$ be a positive kernel as in Definition~\ref{def:positivekernel}. Then
\begin{enumerate}
\item The kernel $\mathbbm{h}: \Xcal \times \Xcal \to \R$ defined by
\begin{equation}
\mathbbm{h}(x,y) = \mathbbm{k}(x,x)+\mathbbm{k}(y,y)-2\mathbbm{k}(x,y) 
\label{eq:definitionhk_general}
\end{equation}
is a negative kernel in the sense of Definition~\ref{def:negativekernel}.
\item For any $z \in \Xcal$ the kernel $\mathbbm{k}_z: \Xcal \times \Xcal \to \R$ defined by
\begin{equation}
	\mathbbm{k}_z(x,y) =  \frac{\mathbbm{h}(x,z)+\mathbbm{h}(y,z)-\mathbbm{h}(x,y)}{2} 
\label{eq:definitionkz}
\end{equation}
is a positive kernel in the sense of Definition~\ref{def:positivekernel}.
\item The kernel $\mathbbm{h}$ is bounded if and only if 
$\mathbbm{k}$ is bounded.
\item For any fixed $z \in \Xcal$ denote~:
\begin{align}
& \Mcal_{\mathbbm{k}} :=	\left\{ \mu \in \Tcal\Vcal(\Xcal):
	\int_\Xcal \sqrt{\mathbbm{k}(x,x)}|\mu|(dx) < \infty\right\}
\\ &
\Mcal^{\mathbbm{h}}	= \left\{ \mu \in \Tcal\Vcal(\Xcal):
	\int_\Xcal \sqrt{\mathbbm{h}(x,z)}|\mu|(dx) < \infty\right\}.
\end{align}
Then
\begin{align}
 \Mcal_{\mathbbm{k}} = \Mcal^{\mathbbm{h}}.
\end{align}
\item \label{item:distance_same_kktilde}
In particular if two positive definite kernels $\mathbbm{k}$ and 
	$\tilde{\mathbbm{k}}$ give same $\mathbbm{h}$ by 
\eqref{eq:definitionhk_general} then~: 
\begin{align}
 \Mcal_{\mathbbm{k}} = \Mcal_{\tilde{\mathbbm{k}}}.
\label{eq:kktilde_same_space}
\end{align}
Moreover for $\eta_1,\eta_2 \in  \Mcal_{\mathbbm{k}} = \Mcal_{\tilde{\mathbbm{k}}}$ with $\int_\Xcal \eta_1(dx) =\int_\Xcal \eta_2(dx) $ we have~:
\begin{align}
\int_\Xcal \int_\Xcal  \mathbbm{k}(x,y) (\eta_1-\eta_2)(dx) (\eta_1-\eta_2)(dy)
=
\int_\Xcal \int_\Xcal  \tilde{\mathbbm{k}}(x,y) (\eta_1-\eta_2)(dx) (\eta_1-\eta_2)(dy) < \infty.
\label{eq:defdistance_kktilde_same}
\end{align}
This says, see \eqref{eq:defdistance_k}, that 
when  $\eta_1$ and $\eta_2$  have the same total mass
the distance between them depends  only on $\mathbbm{h}$ and not on the specific choice of kernel $\mathbbm{k}$.
\end{enumerate}	
\label{lemma:properties_kernels}
\end{lemma}
\begin{proof} {\bf Item 1~:} 
The  conclusion results directly by checking the definition, see also 
 \cite[Property 21.5.4 p. 529]{book_dist_probas13}.
\\ \noindent {\bf Item 2~:} 
We follow \cite[Chapter 3, Lemma 2.1]{berg1984harmonic}
and  
 note first that, using the above relations and after replacing \eqref{eq:definitionhk_general}  in \eqref{eq:definitionkz} one obtains
\begin{equation}
	\mathbbm{k}_z(x,y) =  \mathbbm{k}(x,y)+\mathbbm{k}(z,z)- \mathbbm{k}(y,z)-\mathbbm{k}(x,z). 
\end{equation}
If we are now to check the positivity of 
$	\mathbbm{k}_z(x,y)$ by the Definition~\ref{def:positivekernel} we have to prove that 
$\sum_{j_1,j_2} \alpha_{j_1} \alpha_{j_2} \mathbbm{k}_z(X_{j_1},X_{j_2}) \ge 0$. But this equals 
$$\sum_{j_1,j_2} \alpha_{j_1} \alpha_{j_2} \left[\mathbbm{k}(X_{j_1},X_{j_2})+\mathbbm{k}(z,z)
-\mathbbm{k}(X_{j_1},z) - \mathbbm{k}(z,X_{j_2}) \right]$$
which is positive by using the Definition~\ref{def:positivekernel} for $\mathbbm{k}$, the $J+1$ points
$X_1, ..., X_J, z$ and $J+1$ weights $\alpha_1, ..., \alpha_J, - \sum_j \alpha_j$.
\\ \noindent {\bf Item 3~:} 
When  $\mathbbm{k}$ is bounded it is obvious that   $\mathbbm{h}$ is also bounded. Assume now  $\mathbbm{h}$ is bounded by some constant $C^B$ and prove that  $\mathbbm{k}$
is bounded. If the set $\{\mathbbm{k}(x,x), x \in \Xcal \}$ is bounded, by the inequality 
 \eqref{eq:inequality_kxy_kxxkyy}
%$|\mathbbm{k}(x,y)| \le \sqrt{\mathbbm{k}(x,x)\mathbbm{k}(y,y)}$ 
it follows that $\mathbbm{k}$ is bounded and the conclusion follows. 
\\ 
Let us analyze the situation when 
the set $\{\mathbbm{k}(x,x), x \in \Xcal \}$ is not bounded.
Take a sequence $x_n$ such that $\mathbbm{k}(x_n,x_n) \to \infty$;
since  $\mathbbm{h}$ is bounded $h(x_n,0) \le C^B$, which means that 
$\mathbbm{k}(x_n,x_n) -2 \mathbbm{k}(x_n,0) + \mathbbm{k}(0,0)\le C^B $ so
  $\mathbbm{k}(x_n,0) \to \infty$.
  On the other hand, testing the positivity of  $\mathbbm{k}$ for points $x_n$, $0$ and weights $1,-2$ we have that 
  $ \mathbbm{k}(x_n,x_n)- 4 \mathbbm{k}(x_n,0) + 4\mathbbm{k}(0,0) \ge 0$. But this is not possible because
  $\mathbbm{k}(x_n,x_n)- 4 \mathbbm{k}(x_n,0) + 4\mathbbm{k}(0,0) = 
  \mathbbm{k}(x_n,x_n)- 2 \mathbbm{k}(x_n,0) + \mathbbm{k}(0,0)  )- 2 \mathbbm{k}(x_n,0) + 3\mathbbm{k}(0,0) = \mathbbm{h}(x_n,0)- 2 \mathbbm{k}(x_n,0) + 3\mathbbm{k}(0,0) \to -\infty$ thus we obtained a contradiction.   Therefore
the set $\{\mathbbm{k}(x,x), x \in \Xcal \}$ cannot be unbounded so we are back to previous situation and the assertion is proved.
\\ \noindent {\bf Item 4~:} 
We obtain the conclusion by using the following inequalities:
\begin{align}
	& \forall x,z \in \Xcal~:~ \mathbbm{h}(x,z) \le 2 (\mathbbm{k}(x,x)+ \mathbbm{k}(z,z)),
	\\
	& \forall x,z \in \Xcal~:~ \mathbbm{k}(x,x) \le 2 (\mathbbm{h}(x,z)+ \mathbbm{k}(z,z)).
\end{align}
Both inequalities follow from \eqref{eq:inequality_kxy_kxxkyy} and \eqref{eq:definitionhk_general}; the first is immediate. For the second we write~:
\begin{align}
	& 2 (\mathbbm{k}(x,x) + \mathbbm{k}(z,z)- \mathbbm{h}(x,z)) = 
	4 \mathbbm{k}(x,z) \le 
	4 \sqrt{\mathbbm{k}(x,x) \cdot \mathbbm{k}(z,z)} \le 
	\mathbbm{k}(x,x) + 4\mathbbm{k}(z,z), 
\end{align}
and relation follows by inspecting the first and last terms.
\\ \noindent {\bf Item \ref{item:distance_same_kktilde}~:} The first conclusion \eqref{eq:kktilde_same_space}  is  a mere consequence of the previous item. The second one is more technical because of potential integrability problems. If $\mathbbm{k}$ and 
$\tilde{\mathbbm{k}}$ have same $\mathbbm{h}$ this means that 
\begin{align}
\forall x,y \in \Xcal~: 
\mathbbm{k}(x,x) + \mathbbm{k}(y,y)- 2\mathbbm{k}(x,y) =
 \mathbbm{h}(x,y) = 
 \tilde{\mathbbm{k}}(x,x) + \tilde{\mathbbm{k}}(y,y)- 2\tilde{\mathbbm{k}}(x,y)
\end{align}
  thus
\begin{align}
\tilde{\mathbbm{k}}(x,y)= {\mathbbm{k}}(x,y) + \frac{g(x)+g(y)}{2}, 
\text{ where } g(x) = \tilde{\mathbbm{k}}(x,x)- {\mathbbm{k}}(x,x).
\label{eq:relationkktilde}
\end{align}
An important estimation is that $g(x)$ is absolutely integrable with respect to $\eta_i$. Indeed, take $y=0$ in \eqref{eq:relationkktilde} then~:
%\begin{align}
$
g(x) = 	2\tilde{\mathbbm{k}}(x,0)- 2{\mathbbm{k}}(x,0) - g(0)
$. 
%	\label{eq:ppty_gx}\end{align}
Using \eqref{eq:inequality_kxy_kxxkyy} we obtain $|g(x)|\le c_1+c_2 \left(\sqrt{\mathbbm{k}(x,x)}+\sqrt{\tilde{\mathbbm{k}}(x,x)} \right)$ for some positive constants $c_1$ and $c_2$. Since $\eta_i \in \Mcal_{{\mathbbm{k}}} = \Mcal_{\tilde{\mathbbm{k}}}$ we obtain 
$\int_{\Xcal} |g(x)| |\eta_i|(dx) < \infty$. From here computations are straightforward because all integrals are finite : 
\begin{align}
&	\int_\Xcal \int_\Xcal  \tilde{\mathbbm{k}}(x,y) (\eta_1-\eta_2)(dx) (\eta_1-\eta_2)(dy) = 
\int_\Xcal \int_\Xcal  \mathbbm{k}(x,y) (\eta_1-\eta_2)(dx) (\eta_1-\eta_2)(dy)
\nonumber \\ &	
+	\int_\Xcal \int_\Xcal  \frac{g(x)+g(y)}{2} (\eta_1-\eta_2)(dx) (\eta_1-\eta_2)(dy) 
\nonumber \\ &	
=	\int_\Xcal \int_\Xcal  \mathbbm{k}(x,y) (\eta_1-\eta_2)(dx) (\eta_1-\eta_2)(dy)
+2 \cdot  \int_\Xcal \frac{g(x)}{2} (\eta_1-\eta_2)(dx) \cdot \underbrace{\int_\Xcal (\eta_1-\eta_2)(dy)}_{=0 \text{ by hypothesis}}
\nonumber \\ &	
=	\int_\Xcal \int_\Xcal  \mathbbm{k}(x,y) (\eta_1-\eta_2)(dx) (\eta_1-\eta_2)(dy).
\end{align}
\end{proof}
Any strictly\footnote{When the kernel is only positive definite the same can be proven but the associated Hilbert space is in the form of a quotient.} positive definite kernel $k$ defines, by the Moore-Aronszajn theorem~\cite{aronszajn50}, a unique Hilbert space $\Mcal\Acal_k$ of functions on $\Xcal$ for which $k$ is a reproducing kernel, i.e., $\forall f \in \Mcal\Acal_k : \langle f, k_{x} \rangle_{\Mcal\Acal_k} = f(x)$ where
$k_{x}= k(x,\cdot) \in \Mcal\Acal_k$. This is equivalent to say that for any  $x\in \Xcal$ the evaluation functional 
$L_x : f \in \Mcal\Acal_k \mapsto f(x)=L_x(f)$
is continuous. 
%The special function $k_x$ is called a feature.

The norm in $\Mcal\Acal_k$ of the element $k_x \in \Mcal\Acal_k$ is $k(x,x)$ because in fact 
$\langle k_x, k_y \rangle_{\Mcal\Acal_k} = k(x,y)$.
%Define~:
%\begin{eqnarray}
%%& \ & 
%\Mcal_k = \left\{ \mu \in \Tcal\Vcal(\Xcal)~:
%%\right. \nonumber \\ & \ & \left. 
%\int_\Xcal \sqrt{k(x,x)}|\mu|(dx) < \infty\right\}.
%\label{eq:definition_space_finite_measures_kernel}
%\end{eqnarray}
Recalling the Definition 
\eqref{eq:definition_space_finite_measures_kernel}, 
for any $\mu \in \Mcal_k$ we denote $k_\mu = \int_\Xcal k_x \mu(dx) \in \Mcal\Acal_k$; the following relations hold~:
\begin{equation}
\langle k_\mu, k_\nu \rangle_{\Mcal\Acal_k} = \int_{\Xcal \times \Xcal} k(x,y) \mu(dx)\nu(dy), \ 
\| k_\mu\|_{\Mcal\Acal_k}^2 = \int_{\Xcal \times \Xcal} k(x,y) \mu(dx)\mu(dy).
\label{eq:definition_scalar_product_rkhs}
\end{equation}

 We follow 
\cite[chapter 4]{rkhs_book_berlinet_agnan} (see also 
\cite{sriperumbudur2010hilbert}) and introduce the mapping from $\Mcal_k$
to $\Mcal\Acal_k$ by choosing $\mu \mapsto k_\mu$. This mapping induces a Hilbert space structure on $\Mcal_k$ so, with a slight abuse of notation, we will work with the scalar product~: 
\begin{equation}
\forall \mu,\nu \in \Mcal_k~:
\langle \mu, \nu \rangle_{\Mcal_k} = \langle k_\mu, k_\nu \rangle_{\Mcal\Acal_k} = 
 \int_{\Xcal \times \Xcal} k(x,y) \mu(dx)\nu(dy).
\label{eq:definition_scalar_product_measures}
\end{equation}
This scalar product defines a distance and a norm by the usual relation 
\begin{equation}
\| \eta\|_{\Mcal_k}^2 := 
\| k_\eta\|_{\Mcal\Acal_k}^2 = \langle k_\eta,k_\eta \rangle_{\Mcal\Acal_k}= 
\langle \eta,\eta \rangle_{\Mcal_k}.
\label{eq:embedding_norm}
\end{equation}

We have thus embedded the measures in $\Mcal_k$~\footnote{Note that the image $\{k_\mu | \mu \in \Mcal_k\}$ of $\Mcal_k$ through this embedding is not necessarily a Hilbert space itself because it may not be closed under the norm of  $\Mcal\Acal_k$. The technical term for $\Mcal_k$ is ``Hausdorff pre-Hilbert space" because we do not know if it is complete with respect to the topology induced by the norm \eqref{eq:embedding_norm}. For additional details on the Hilbert topology see also
\cite{Guilbart1979}.} in a Hilbert space.
Immediate computations show that the squared distance  $(x,y) \mapsto \|\delta_x - \delta_y\|^2_{\Mcal_k}$ is a negative definite kernel.
Note that the embedding is not expected to be surjective, i.e., there may exist functions in $\Mcal\Acal_k$ that do not correspond to any measure $\xi \in \Mcal_k$.
%For instance when $k$ is the Gaussian kernel in \eqref{eq:gaussian_distance}
%we know that $\Mcal\Acal_k$ contains the constant functions (the kernel being universal, cite \cite{micchelli1984interpolation} but this does not correspond to a measure.
Reciprocally, 
Schoenberg proved in \cite{schoenberg1938metric}
that a metric space $(Y, d)$ can be isometrically embedded into some real Hilbert space if and only if $d^2(\cdot, \dot)$ is a negative definite kernel. 
A characteristic property of metric spaces that can be embedded into a Hilbert space is that the following ``parallelogram identity"\footnote{In fact is can be proved that $L=2$ implies the identity for all other $L>2$.} holds~:
\begin{eqnarray}
	&  \forall \nu_1,\nu_2, \mu, \forall \lambda \in 
	\R~: &  
		\label{eq:parallelogram_general_metric}
	 \\ &   
	d(\mu, \lambda \nu_1 + (1-\lambda)\nu_2)^2 & = 
	\lambda d(\mu,  \nu_1 )^2  +
	(1-\lambda) d(\mu,  \nu_2 )^2  
	- \lambda(1-\lambda) d(\nu_1,  \nu_2 )^2.  \nonumber
\end{eqnarray}
This relation can be readily generalized for more than $2$ measures; for a given $\mu$, taking  $Y= \{ \chi \in \Mcal^h ; \int \chi(dx)= \int \mu(dx) \}$ we obtain~:
\begin{eqnarray}
 &  \ & \forall L \in \N, L \ge 2,~ \nu_1, ..., \nu_L, \mu \in  \Mcal^h,  \text{ with }
 \int \nu_\ell(dx)= \int \mu(dx), \ell \le L~
  \nonumber   \\ &   \ &   \forall \lambda= (\lambda_\ell)\in \R^L \text{ with }\sum_{\ell=1}^L \lambda_\ell=1~: 
  \nonumber   \\ &   \ & 
     d\left(\mu, \sum_{\ell=1}^L \lambda_\ell \nu_\ell \right)^2  = 
    \sum_{\ell=1}^L \lambda_\ell d(\mu,  \nu_\ell )^2  
    - \frac{1}{2} \sum_{\ell,\ell'=1}^L \lambda_\ell \lambda_{\ell'} d(\nu_\ell,  \nu_{\ell'} )^2. 
    \label{eq:parallelogram}\end{eqnarray}

The distance function can be resumed to the knowledge of 
$h(x,y)=d(\delta_x,\delta_y)^2$. When $k$ is given
a useful immediate identity involving $k$ and $h$ is \eqref{eq:relation_h_function_of_k}.
%\begin{equation}
%    \forall x,y \in \Xcal~: h(x,y) = k(x,x)+k(y,y)-2k(x,y).
%\end{equation}
On the other hand when $h$ is given several $k$ can be compatible with the same $h$; a classic example (see~\cite{sriperumbudur2010hilbert}) is to work with~:
\begin{equation}
    k_{z_0}(x,y) = \frac{h(x,z_0)+h(y,z_0)-h(x,y)}{2}, \ 
      x,y \in \Xcal,
\label{eq:relation_h_gives_k_z0}
\end{equation}
where $z_0\in \Xcal$ is arbitrary (but fixed). The Hilbert space 
$\Mcal\Acal_{ k_{z_0}}$
associated to  $k_{z_0}$ by the Moore-Aronszajn theorem
is the same for all $z_0$.

\newpage

\section{Proofs and additional remarks}
\subsection{Proof of Lemma~\ref{lem:lowerbound_implies_coercive}}
\label{sec:proof_lemma_lowerbound_implies_coercive}

\begin{proof} 
Since $h$ satisfies \eqref{eq:relation_h_function_of_k} then 
 $\forall x \in \Xcal : $ $h(x,x)=0$. 
We work in the Hilbert embedding induced by the kernel
\begin{equation}
k_0(x,y) = \frac{h(x,0) + h(y,0) - h(x,y)}{2},
\label{eq:relation_h_function_of_k_cst_is_0}
\end{equation}
which also satisfies \eqref{eq:relation_h_function_of_k}. 
In particular 
$h(x)=d(\delta_x,\delta_0)^2=h(x,0)= k_0(x,x) = \langle \delta_x,\delta_x \rangle = \|\delta_x\|^2$. 
First note that assumption \eqref{eq:hyp_lowerbound_h} shows that, for any $x,y\in \Xcal$:
$2 \langle \delta_x, \delta_y \rangle = 
\|\delta_x\|^2 + \|\delta_x\|^2 - \|\delta_x-\delta_y\|^2 = h(x)+h(y)-h(x,y) \ge C_L$ thus, denoting $C_p=C_L/2$~:
\begin{equation}
\forall x,y \in \Xcal : \ \langle \delta_x, \delta_y \rangle  \ge  C_p.
\label{eq:lowerbound}
\end{equation}
Hence
\begin{eqnarray}
& \ & 	
  \left\| \sum_{j=1}^J \beta_j \delta_{X_j}\right\|^2 =
\sum_{j=1}^J (\beta_j)^2 \| \delta_{X_j}\|^2 + 
\sum_{j,q=1, j \neq q }^J \beta_j \beta_q
\langle \delta_{X_j} , \delta_{X_q} \rangle
\nonumber \\ & \ & 
\ge  \sum_{j=1}^J (\beta_j)^2 h(X_j)
+ C_p \left(1- \sum_j (\beta_j)^2\right)
\ge  \sum_{j=1}^J (\beta_j)^2 h(X_j) -|C_p|,
\label{eq:estimation_bounded_alpha_X}
\end{eqnarray}
where we used the relation \eqref{eq:lowerbound} and
the fact that $\beta_j$ are positive and sum up to one.

But, since by hypothesis $h$ tends to $+\infty$ at infinity
we obtain 
$\lim_{X \to \infty}   \left\| \sum_{j=1}^J \beta_j \delta_{X_j}\right\|^2 = \infty$ and thus~:
\begin{equation}
\lim_{X \to \infty}  d \left( 
\sum_{j=1}^J \beta_j \delta_{X_j}, \delta_0\right) \ge  
\lim_{X \to \infty}  
\left\| \sum_{j=1}^J \beta_j \delta_{X_j}\right\| - \|\delta_0\| = \infty,
\end{equation}
which proves the first conclusion. The conclusion for the particular kernels is obtained by straightforward computations because both satisfy hypothesis of the lemma. 
\end{proof}

\subsection{Proof of Lemma~\ref{lemma:dist_r_inequalities}}
\label{sec:proof_lemma_dist_r_inequalities}

\begin{proof}
We will only prove the assertion when $\eta$, $\mu$ are probability measures, the extension to finite total variation being a simple consequence of the additive and multiplicative properties of the distance (because of the Hilbert space embedding). Note that the requirement $\int (\eta-\mu)(dx) =0$ is not source of particular technical problems but the extension to $\int (\eta-\mu)(dx) \neq 0$ 
is not necessary in the following.

Recall that for $r\in]0,1[$ and $t\ge 0$ :  
$t^r= \frac{1}{-\Gamma(-r)} \int_0^\infty \frac{1-e^{-ts}}{s^{r+1}}ds$ 
 where $\Gamma(\cdot)$ is the Euler gamma function 
(see for instance \cite{schoenberg1938metric} and Corollary \ref{cor:huber_energy_distance_coercive} below) ; 
 for $t=\|x-y\|^2/2$ we obtain 
  $\dfrak_r(\delta_x,\delta_y)^2=\|x-y\|^{2r}= \frac{1}{-\Gamma(-r)}  \int_0^\infty \frac{1-e^{-s\|x-y\|^2/2}}{s^{r+1}}ds = 
 \frac{1}{-\Gamma(-r)}  \int_0^\infty \frac{\gfrak_{1/\sqrt{s}}(\delta_x,\delta_y)^2}{s^{r+1}}ds$. Thus for any $r\in ]0,2[$ and some constant 
$C''_r>0$~:
 \begin{equation}
 \dfrak_r(\eta_1,\eta_2)^2 =  C''_r \int_0^\infty \frac{\gfrak_{1/\sqrt{s}}(\eta_1,\eta_2)^2}{s^{r+1}}ds,
\ \forall \eta_i \in \Pcal(\Xcal) \text{ with }   \dfrak_r(\eta_i,\delta_0) < \infty, i=1,2.
 \end{equation}
 Note that since the Gaussian kernel is bounded, any distance among probability distributions is bounded by some fixed constant and thus in the formula above the part
 $\int_1^\infty \frac{\gfrak_{1/\sqrt{s}}(\eta_1,\eta_2)^2}{s^{r+1}}ds$ is bounded by some constant depending on $r$.
 On the other hand for $0<r'<r$: 
 $\int_0^1 \frac{\gfrak_{1/\sqrt{s}}(\eta_1,\eta_2)^2}{s^{r'+1}}ds \le \int_0^1 \frac{\gfrak_{1/\sqrt{s}}(\eta_1,\eta_2)^2}{s^{r+1}}ds$. Combining the two bounds we obtain the conclusion.
 \end{proof}

%%%%%%%%%%%%%
\subsection{Proof of Corollary~\ref{cor:huber_energy_distance_coercive}}
\label{sec:proof_cor_huber_energy_distance_coercive}

\begin{proof}
The conclusion follows exactly the same path as in the proof of the
 Lemma~\ref{lemma:dist_r_coercive} 
if we
make use of the formula~:
\begin{equation}
(a+t)^r-a^r = \frac{1}{-\Gamma(-r)} \int_0^\infty \frac{(1-e^{ts})e^{-as}}{s^{1+r}} ds 
\text{ for } r \in [0,1[,a,t \ge 0.
\label{eq:mixture_formula_sqrtHE}
\end{equation}
%equation for the reverse one : 
%check with https://www.wolframalpha.com/input?key=&i=inverse+Laplace+transform+%28d%2Fdt%28t%5E%7B2*r%7D%2F%28a%2Bt%29%5E%7Br%7D%29

Note that in fact the relation
\eqref{eq:estimation_distancesE}
in Lemma~\ref{lemma:dist_r_inequalities} extends to the class of distances $\dfrak_{r,a}^{HE}$ 
%and the class $\dfrak_{r,a}^{RHE}$ 
with $a$ fixed and $r$ variable.

For completeness we prove \eqref{eq:mixture_formula_sqrtHE}; a short analysis shows that the integral is indeed well defined (finite) near $s=0$ and $s=\infty$; we write~:  
\begin{eqnarray}
& \ & 
\int_0^\infty \frac{(1-e^{ts})e^{-as}}{s^{1+r}} ds 
=
\int_0^\infty \frac{e^{-as}- e^{-(a+t)s}}{s^{1+r}} ds 
= \int_0^\infty \int_{a}^{a+t} s e^{-u s}du 
\frac{ds}{s^{1+r}} 
\nonumber \\ & \ & 
= \int_{a}^{a+t}
\int_0^\infty  e^{-u s} s^{-r} ds du =
\int_{a}^{a+t} u^{r-1}
\int_0^\infty e^{-w} w^{-r} dw du 
\nonumber \\ & \ & 
=\Gamma(1-r) \left. \frac{u^r}{r} \right|^{a+t}_a 
= -\Gamma(-r) [(a+t)^r-a^r].
\end{eqnarray}
\end{proof}
%

%%%%%%%%%%%%%%%%%%%%%%%%%%%%%%%%%%%%
\subsection{Proof of Proposition~\ref{prop:existence_fixed_alpha}}
\label{sec:proof_prop:existence_fixed_alph}

\begin{proof}
To fix the constants and ease the notation we can consider
that $\int_\Xcal \eta(dx) = \sum_q \alpha_q=1$ otherwise replace
$\delta_0 $ by $ \delta_0 \cdot \sum_q \alpha_q $ in all that follows.
Let us denote 
$f(X):= d \left(\delta_{\alpha,X} ,\eta \right)^2$ and 
$m_\eta$ the infimum in \eqref{eq:minimization}.
Take a sequence $(X^n)_{n\ge 1}$ such that 
$f(X^n) \to m_\eta$. The strategy of the proof is to show that we can extract a converging sub-sequence which has a finite limit and whose distance to the $\eta$ converges to $m_\eta$. We can suppose without any loss of generality that  $f(X^n)\le m_\eta + 1$. 
Then :
\begin{equation}
 m_\eta + 1 \ge f(X^n) =  d \left(\delta_{\alpha,X^n} ,\eta \right)^2 \ge 
 \frac{d \left(\delta_{\alpha,X^n} ,\delta_0 \right)^2 -2 d \left(\delta_0 ,\eta \right)^2 }{2}, 
\end{equation}
which implies 
\begin{equation}
d \left( \delta_{\alpha,X^n},\delta_0 \right) ^2 \le 
2(	m_\eta + 1) + 2 d \left(\delta_0 ,\eta \right)^2 < \infty.%2  \|\eta \|^2.
\label{eq:estimate_bounded_norm}
\end{equation}
Since the kernel $h$ is measure coercive, the sequence $X_q^n$ must be bounded. We can extract converging subsequences (we keep the same notation for the indices) and let $X^\star := \lim_{n \to \infty} X^n$. Note that $X^n_q \to X_q^\star$ implies 
$\|\delta_{X_q^n}- \delta_{X_q^\star}\|^2 = h(X_q^n,X_q^\star) \to h(X_q^\star,X_q^\star)=0$ thus $\delta_{X_q^n}\to \delta_{X_q^\star}$ (we used the continuity of $h$)~; furthermore, 
$\delta_{\alpha,X^n} = \sum_q \alpha_q \delta_{X_q^n} \to \delta_{\alpha,X^\star}$ and by the continuity of the distance
 we obtain that $m_\eta = \lim_n f(X^n) = f(X^\star)$ which means that $X^\star$ is a solution of the minimization problem \eqref{eq:minimization}. 
\end{proof}

%%%%%%%%%%%%%%%%%%%%%%%%%%%%%%%%%
\subsection{Proof of Proposition~\ref{prop:existence_variable_alpha}}
\label{sec:proof_prop:existence_variable_alpha}

\begin{proof}
First remark that~\eqref{eq:lower_bound_kernel_proof_exitstence_variable_weights}
implies that $h$ is measure coercive; denote $m_\eta$ the minimum in 
\eqref{eq:minimization_variable_weights}.
Consider $\alpha^n, X^n$ a minimizing sequence; of course, the norm of 
$\delta_{\alpha^n,X^n}$ is finite (same arguments as in Proposition~\ref{prop:existence_fixed_alpha} estimation  \eqref{eq:estimate_bounded_norm}). Working as in the proof of the
Lemma~\ref{lem:lowerbound_implies_coercive}
% lemma~\eqref{lem:lowerbound_implies_coercive} 
estimation \eqref{eq:estimation_bounded_alpha_X}
(recall that $\|\delta_x\|^2= k(x,x)$) 
we obtain that $\| \delta_{\alpha^n,X^n} \|^2$ is lower bounded by 
$\sum_q \alpha_q^n k(X^n_q,X^n_q)$ which shows that 
the sequences $n \mapsto \alpha_q^n k(X^n_q,X^n_q)$ are bounded for any 
$q \le Q$. 
On the other hand, since $\alpha_q^n$ are all positive and of prescribed total sum, they belong to a compact space and there is a sub-sequence that converges to some $\alpha^\star$. Proceeding sequentially (we renote the resulting sub-sequence with the index ``n"), one ends up with a partition $\Bcal \cap \Ucal$ of $\{1,...,Q\}$ such that~: 

- for any $q \in \Bcal$ the sequence $X_q^n$ converges to some finite value
$X_q^\star$; in this case $\lim_{n\to\infty}\alpha^n_q\delta_{X^n_q} =
\alpha^\star\delta_{X^\star}$ in the sense of strong convergence in $\Mcal\Acal_k$;
denote $\xi^b = \sum_{q \in \Bcal} \alpha^\star\delta_{X^\star}$.

- for any $q \in \Ucal$~:~$X_q^n\to \infty$ and in this case necessarily $\lim_{n\to\infty}\alpha^n_q=0$, and moreover 
 $\lim_{n\to\infty}\alpha^n_q k(X^n_q,X^n_q)$ is bounded.

Consider $q\in\Ucal$; we will prove that $\alpha_q^n \delta_{X^n_q}$ converges weakly to zero in $\Mcal\Acal_k$ (where we used the embedding 
introduced in Appendix \ref{sec:appendix_embedding}). We can suppose, without loos of generality, that $\alpha_q^n$ are non-null from some $n$ forward (otherwise consider the sub-sequence where $\alpha_q^n$ are all null and the convergence to zero is attained). write   
$\alpha_q^n \delta_{X^n_q} =\left(\alpha_q^n \cdot \|\delta_{X^n_q}\|\right)  \cdot \frac{\delta_{X^n_q}}{\|\delta_{X^n_q}\|}$; 
in particular note that $\alpha_q^n \cdot \|\delta_{X^n_q}\|$
must be bounded. 
The sequence of general term
$\frac{\delta_{X^n_q}}{\|\delta_{X^n_q}\|}$ is bounded thus in $\Mcal\Acal_k$
it converges to some $\xi \in \Mcal\Acal_k$ of norm at most $1$. On the other hand, for any $y \in \Xcal$~:
\begin{equation}
    \lim_{n\to \infty} 
    \left\langle \delta_y, \frac{\delta_{X^n_q}}{\|\delta_{X^n_q}\|} \right\rangle
=   \frac{k(y,X^n_q)}{\sqrt{k({X^n_q},{X^n_q})}} \to 0,
\end{equation}
where we used 
\eqref{eq:assumption_norm_kx_infinite} and \eqref{eq:assumption_bounded_scalar_product} and the fact that 
$x^n_q \to \infty$. But since this is true for any $y$, we obtain that $\xi=0$.
Since in addition $\alpha_q^n \cdot \|\delta_{X^n_q}\|$
is bounded for any $q \in \Ucal$ 
it follows that~$\sum_{q \in \Ucal} \alpha_q^n \delta_{X^n_q}$ converges weakly to zero. Since on the other hand 
 $\eta-\sum_{q \in \Ucal} \alpha_q^n \delta_{X^n_q}$ converges strongly to $\eta - \xi^b$ we obtain~:
 \begin{equation}
\lim_{n\to \infty}
 \left\langle 
 \sum_{q \in \Ucal} \alpha_q^n \delta_{X^n_q},
\eta- \sum_{q \in \Bcal} \alpha_q^n \delta_{X^n_q}
 \right\rangle    = 0.
 \end{equation}
We can write~:
\begin{eqnarray}
& \ &     m_\eta  = \lim_{n\to \infty}
\left\| \eta- \sum_{q=1}^Q \alpha_q^n \delta_{X^n_q}
\right\|^2 = 
\lim_{n\to \infty} \left\| \eta- \sum_{q\in \Bcal} \alpha_q^n \delta_{X^n_q}
\right\|^2  + 
\lim_{n\to \infty} \left\|\sum_{q\in \Ucal} \alpha_q^n \delta_{X^n_q}
\right\|^2 
\nonumber \\ & \ & 
- 2  \left\langle 
 \sum_{q \in \Ucal} \alpha_q^n \delta_{X^n_q},
\eta- \sum_{q \in \Bcal} \alpha_q^n \delta_{X^n_q}
 \right\rangle
= \| \eta- \xi^b \|^2  + 
\lim_{n\to \infty} \left\|\sum_{q\in \Ucal} \alpha_q^n \delta_{X^n_q}
\right\|^2
\nonumber \\ & \ & 
\ge \| \eta- \xi^b \|^2.
\end{eqnarray}
But, on the other hand, $\xi^b$ is an admissible candidate for the problem 
\eqref{eq:minimization_variable_weights} which means that 
$\| \eta- \xi^b \|^2 \le m_\eta$; so ultimately 
$\| \eta- \xi^b \|^2 = m_\eta$, thus $\xi^b$ is a solution of~\eqref{eq:minimization_variable_weights}.
\end{proof}

%%%%%%%%%%%%%%%%%%%%%%%%%%%%
\subsection{Proof of the Proposition~\ref{prop:existence_gaussian_kernel}}
\label{sec:proof_prop:existence_gaussian_kernel}

\begin{proof}
Without loss of generality we can set the constant $a$ equal to $1$ and denote $\gfrak=\gfrak_a$; also denote $f(\alpha,X) = \gfrak \left(\delta_{\alpha,X} ,\eta \right)^2$. When $\alpha$ is fixed we will only write the $X$ argument.

We start with the proof of the \eqref{eq:min_problem_gaussian_variable_alpha} which is more difficult. %Its difficulty is in proving that some $\alphaq$ will not vanish while some $X_q$ can tend to infinity. 
Consider thus a minimizing sequence i.e., 
$(\alpha^n, X^n)_{n \ge 1}$ such that  $f(\alpha^n, X^n) \to m_\eta$, $m_\eta$ being the minimum value in \eqref{eq:min_problem_gaussian_variable_alpha} (we know it is finite because is positive and bounded by $\gfrak \left(\delta_{(1,0,...),{\bf 0}} ,\eta \right)^2$). For any coordinate $q \le Q$ such that  $X^n_q$ has a bounded sub-sequence we extract a converging sub-sequence. We also can extract converging sub-sequences of the (bounded) sequence $\alpha^n$; to ease notations we renote the resulting sub-sequence with the index $n$ too; we are thus left  with the following situation: set of indices $\{1,2, ... , Q\}$ is partitioned in two  :
 a part $\Bcal$ that we will call ``bounded" and a part $\Ucal$ that we will call ``unbounded"
 such that for some $\alpha^\dag \in \Pcal_Q$~:
\begin{eqnarray}
& \ & 
\forall q\in \Bcal~:~\lim_{n \to \infty}\alpha^n_q = \alpha^\dag_q, 
\lim_{n \to \infty}X^n_q = X^\dag_q \in \R \\ & \ & 
\forall q\in \Ucal~:~\lim_{n \to \infty}\alpha^n_q = \alpha^\dag_q,
\lim_{n \to \infty}X^n_q = \infty.
\end{eqnarray} 
We consider the embedding Hilbert space $\Mcal\Acal_g$ having $g$ as scalar-product i.e., $\langle \delta_x, \delta_y \rangle = g(x,y)$, see Appendix~\ref{sec:appendix_embedding}. 
Note that because of the definition of the $\gfrak$ 
the measure 
$\sum_{q \in \Bcal} \alpha^n_q \delta_{X^n_q}$ converges strongly (i.e. in distance) when $n\to \infty$  to 
$\sum_{q \in \Bcal} \alpha^\dag_q \delta_{X^\dag_q}$ that we will denote $\xi^b$.
If the total mass $z = \int_{\R^N} \xi^b(dx) = \sum_{q \in \Bcal} \alpha^\dag_q$ is equal to $1$ then 
$\Ucal=\emptyset$ and 
the proof is complete. Otherwise suppose $z<1$.

Since $ \langle g(x,\cdot), g(y,\cdot) \rangle_{\Mcal\Acal_g} = g(x,y)$ when $x$ is fixed (or converges to a finite value) and $y\to \infty$ we obtain  $g(x-y)\to 0$. But since $x$ was arbitrary, this means that in  $\Mcal\Acal_g$ the sequence $g(y,\cdot)$ weakly converges to zero when $y\to\infty$\footnote{We use the fact that 
$\Mcal\Acal_g$ is the completion of the
the linear space of functions $g(x,\cdot)$ for $x\in \R^N$.}.
Therefore, for any $q \in \Ucal$ any cross scalar product of the type~:
%\begin{eqnarray}
$ \left\langle \sum_{q \in \Bcal} \alpha^n_q \delta_{X^n_q} - \eta, \alpha^n_q \delta_{X^n_q}
\right\rangle
$
%\label{eq:unbounded_perpendicular0}\end{eqnarray}
converges to zero when $n\to \infty$. We can  write~:
\begin{eqnarray} & \ & 
m_\eta =\lim_{n\to \infty} f(\alpha_n,X^n) = 
\lim_{n\to \infty}
\left\| \sum_{q \in \Bcal} \alpha^n_q \delta_{X^n_q}  + 
\sum_{q \in \Ucal} \alpha^n_q \delta_{X^n_q} - \eta \right\|^2 
\nonumber \\ & \ & 
=
\lim_{n\to \infty}
\left\| \sum_{q \in \Bcal} \alpha^n_q \delta_{X^n_q} 
 - \eta \right\|^2 + 0 + 
\lim_{n\to \infty}
\left\| \sum_{q \in \Ucal} \alpha^n_q \delta_{X^n_q} \right\|^2 
%\nonumber \\ & \ & 
\ge 
\left\| \xi^b - \eta \right\|^2 +  
\sum_{q \in \Ucal} (\alpha^\dag_q)^2.
\label{eq:inequlity_gaussian_contradiction}
\end{eqnarray}

The next step is to find some $x^\star \in \R^N$ such that 
$\langle \delta_{x^\star}, \eta-\xi^b \rangle_{\Mcal^\gfrak} > 0$. To this end note first that 
$\int_{\R^N} 1 \cdot [\eta(dx)-\xi^b(dx)] = 1-z >0$ (recall that when $z=1$ the conclusion is already proved). By the Beppo-Levy monotone convergence theorem (we treat $\eta$ and $\xi^b$  separately) we obtain
$\lim_{a\to \infty} 
\int_{\R^N} e^{- \frac{\|x\|^2}{2 a^2}} \cdot  [\eta(dx)-\xi^b(dx)] = 
\int_{\R^N} 1 \cdot  [\eta(dx)-\xi^b(dx)] >0$. Thus for some $a^\star < \infty$ (that can be taken as large as we want)~: $\int_{\R^N} e^{- \frac{\|x\|^2}{2 {a^\star}^2}} \cdot  [\eta(dx)-\xi^b(dx)] >0$. But~:
\begin{eqnarray} &
0 < \int_{\R^N} e^{- \frac{\|x\|^2}{2 {a^\star}^2}} \cdot  [\eta(dx)-\xi^b(dx)]
= c_1 
\int_{\R^N \times \R^N} 
e^{- \frac{\|y\|^2}{2 {b^\star}^2}}
e^{- \frac{\|x-y\|^2}{2}} \cdot  [\eta(dx)-\xi^b(dx)] dy
& \nonumber \\ &  
= c_1\int_{\R^N \times \R^N}  e^{- \frac{\|y\|^2}{2 {c^\star}^2}}  g(x,y) [\eta(dx)-\xi^b(dx)] dy
 = c_1 \int_{\R^N } e^{- \frac{\|y\|^2}{2 { ^\star}^2}} \langle \delta_y, \eta-\xi^b\rangle  dy.
& \end{eqnarray}
Here $c_1$ and $c^\star$ are constants only depending on $a^\star$.
This means that at least one $x^\star$ exists such that 
$ \langle \delta_{x^\star}, \eta-\xi^b\rangle > 0$.
We choose now an index $q^\star \in \Ucal$ and replace the sequence $X^n_{q^\star}$ by $x^\star$ and 
let all other $X^n_{q}$ converge to infinity with requirement that all distances  $ \| X^n_{q}- X^{n}_{q'}\|$ 
also converge to 
$\infty$ as soon as $q \neq q'$, $q,q' \in \Ucal$; this means that 
$\langle \delta_{X^n_{q}}, \delta_{X^{n}_{q'}} \rangle \to 0$ as $n\to \infty$. 
Then, a cumbersome but straightforward computation allows to write~:
\begin{equation}
\lim_{n\to \infty}
\left\| \sum_{q \in \Bcal} \alpha^n_q \delta_{X^n_q}  + 
\sum_{q \in \Ucal} \alpha^n_q \delta_{X^n_q} - \eta \right\|^2 
=
\left\|\eta- \xi^b \right\|^2 +  
\sum_{q \in \Ucal} (\alpha^\dag_q)^2- 2\langle \delta_{x^\star}, \eta-\xi^b\rangle.
\end{equation}
But since the sequences $\alpha^n,X^n$ are admissible
candidates for the minimization problem \eqref{eq:min_problem_gaussian_variable_alpha} 
it follows that 
$m_\eta \le  \left\|\eta- \xi^b \right\|^2 +  
\sum_{q \in \Ucal} (\alpha^\dag_q)^2- 2 \langle  \delta_{x^\star}, \eta-\xi^b\rangle  <  \left\|\eta- \xi^b \right\|^2 +  
\sum_{q \in \Ucal} (\alpha^\dag_q)^2$. We obtained a contradiction with inequality \eqref{eq:inequlity_gaussian_contradiction}. Therefore the assumption $z<1$ is false and thus $z=1$, $\Ucal = \emptyset$ and  $\xi^b$ is a solution of the minimization problem \eqref{eq:min_problem_gaussian_variable_alpha}.

The proof for \eqref{eq:min_problem_gaussian_fixed_alpha}
is a simple repetition of the proof for \eqref{eq:min_problem_gaussian_variable_alpha} but in this case all $\alpha_n$ are constant equal to $\alpha$  and there is no need to extract converging sub-sequences.
\end{proof}

\begin{remark}[existence for the bounded kernel]
The previous proof can be extended to the situation of a more general bounded kernel $k$, provided we keep some important hypothesis as the fact that for any $x$  the function $k(x,\cdot)$ vanishes at infinity or that the diagonal $k(x,x)$ is constant.

On the other hand note that an example similar to Example~\ref{ex:unbounded_non_coercive} can be constructed also for the bounded case that shows that some hypotheses are required in order to obtain existence of a solution.

\noindent For completeness we state the equivalent result for general $\Tcal\Vcal$ measures~:
\begin{corollary}[existence of measure quantization for the Gaussian kernel, $\Tcal\Vcal$ measures] 
	Consider the Gaussian kernel $\gfrak_a$ defined in~\eqref{eq:gaussian_distance}. Let
	$\eta \in \Tcal \Vcal$ with $\int \eta(dx)>0$ and fix an integer $Q \ge 1$.
	\begin{enumerate}
		\item 
		For a given $\alpha\in (\R_+)^Q$ with $\sum_q \alpha_q =\int \eta(dx)$ the minimization problem~:
		\begin{equation}
			\inf_{X=(x_q)_{q=1}^Q \in \R^{N\times Q} } \gfrak_a \left(\delta_{\alpha,X} ,\eta \right)^2 
			\label{eq:min_problem_gaussian_fixed_alphaTV}
		\end{equation}
		admits at least one solution $X^\star \in \R^{N\times Q}$.
		\item 
		The minimization problem~:
		\begin{equation}
			\inf_{X=(x_q)_{q=1}^Q \in \R^{N\times Q}, \ 
				\alpha\in (\R_+)^Q, \sum_q \alpha_q =\int \eta(dx)} \gfrak_a \left(\delta_{\alpha,X} ,\eta \right)^2 
			\label{eq:min_problem_gaussian_variable_alpha_TV}
		\end{equation}
		admits at least one solution $X^\dag$, $\alpha^\dag$.
	\end{enumerate}
	\label{prop:existence_gaussian_kernel_TV}
\end{corollary}
\begin{proof}
	The proof is the exact analog, for general $\Tcal\Vcal$ measures, of the proof above.
\end{proof}

\end{remark}

\subsection{Proof of the Proposition~\ref{prop:blue_estimator}} \label{sec:proof_prop:blue_estimator}

\begin{proof}
\noindent {\bf Item~\ref{item:blue2samples}:} 
We will use the compact writing involving the total sample $X$ and the matrix $w$. Note first that the hypothesis do not allow to use the Gauss-Markov theorem because the law is not the same for all indices. Also note that the values of $w_{a,a}$ are irrelevant because they multiply $d^2(\delta_{X_a},\delta_{X_a})=0$. For the rest of the proof we set $w_{a,a}=0$ for all
$a \le Q+J$.

Let us compute the expectation of the estimator $\widehat{d^2}^w$ for a general matrix $w$. 
\begin{eqnarray}
& \ & 
\Ebb[\widehat{d^2}^w] = \Ebb \left[\sum_{a,b \le Q+J}w_{a,b} d^2(\delta_{X_a},\delta_{X_b})
 \right] = 
\left(\sum_{a,b \le Q, a\neq b} w_{a,b} \right) 
\Ebb_{\substack{X,X' \sim \nu \\ X \independent X'}} \left[d^2(\delta_{X},\delta_{X'}) \right]
\nonumber \\ & \ &
+ \left(\sum_{a,b > Q, a\neq b} w_{a,b} \right) 
\Ebb_{\substack{Y,Y' \sim \mu\\ Y \independent Y'}} \left[d^2(\delta_{Y},\delta_{Y'}) \right]
%\nonumber \\ & \ &
+ \left(\sum_{a \le Q< b} w_{a,b} + w_{b,a} \right)
\Ebb_{\substack{X \sim \nu \\ Y \sim \mu \\ X \independent Y}}
 \left[d^2(\delta_{X},\delta_{Y}) \right]. 
\end{eqnarray}
But on the other hand \eqref{eq:defdistance} can be written as (cf.~\cite[eqn. (2.3)]{distance_based_ann_stat2013})~:
\begin{equation}
d^2(\nu,\mu) =  
\Ebb_{\substack{X \sim \nu \\ Y \sim \mu \\ X \independent Y}} \left[d^2(\delta_{X},\delta_{Y}) \right] -\frac{1}{2}
\Ebb_{\substack{X,X' \sim \nu \\ X \independent X'}} \left[d^2(\delta_{X},\delta_{X'}) \right]
-\frac{1}{2}
\Ebb_{\substack{Y,Y' \sim \mu\\ Y \independent Y'}} \left[d^2(\delta_{Y},\delta_{Y'}) \right]
\end{equation}
We conclude that the estimator $\widehat{d^2}^w$ is unbiased if and only if~:
\begin{equation}
%\left(
\sum_{a,b \le Q, a\neq b} w_{a,b}
% \right) 
=-1/2, 
% \left(
\sum_{a,b > Q, a\neq b} w_{a,b} 
%\right)
=-1/2,  
% \left(
\sum_{a \le Q< b} w_{a,b} +w_{b,a}
%\right)
=1. \label{eq:constraints_bias_zero_blue}
\end{equation}
Denote now by $f_w$ the variance
 $\mathbb{V}(\widehat{d^2}^w)$
 of the estimator $\widehat{d^2}^w$. 
 If we view $w$ as a  
  vector in $\R^{(Q+J)^2}$ then 
 $f(w) = \langle w, \Sigma w\rangle$ where $\Sigma \in \R^{(Q+J)^2 \times (Q+J)^2}$
 is the covariance matrix of the 
$(Q+J)^2$ variables $d^2(\delta_{X_a},\delta_{X_b})$. The matrix $\Sigma$ is always  positive definite  so the function $f(w)$ is convex. 
Let $\Scal_Q$ be the ensemble of permutations of the indices $1,...,Q$
and  $\Scal_J$ the ensemble of permutations of the indices $Q+1,...,Q+J$.
Since the law of $(X_1,...,X_Q)$ is symmetric and that of 
$(X_{Q+1},...,X_{Q+J})$ too, any statistic involving the estimator $\widehat{d^2}^w$ (and in particular its variance) is invariant with respect to permutations $\pi \in \Scal_Q$
and $\rho \in \Scal_J$. This means that 
$f(w) = f(w_1,...,w_{(Q+J)^2}) = f(w_{\pi(1),\pi(1)},..., 
w_{\rho(Q+J),\rho(Q+J)} ) =:f(w_{\pi \otimes \rho})$ where the last identity is a notation.
The convexity implies that~:
\begin{equation}
f(w) = \frac{1}{Q! J!}	\sum_{\pi \in \Scal_Q, \rho \in \Scal_J} f(w_{\pi \otimes \rho}) \ge f \left( \frac{1}{Q! J!}	\sum_{\pi \in \Scal_Q, \rho \in \Scal_J} w_{\pi \otimes \rho}\right).
\label{eq:blue_iterm1_variance_ineq}
\end{equation}
We switch back to the matrix notation for $w$, which is more comfortable in the following. 
We 
will prove that, for any $w$ which satisfies \eqref{eq:constraints_bias_zero_blue} (recall that we set $w_{a,a}=0$ for all $a$)~: 
\begin{equation}
\label{eq:blue_sum_over_permutations}
\frac{\sum_{\pi \in \Scal_Q, \rho \in \Scal_J} w_{\pi \otimes \rho}}{Q! J!}  
= %w^\star, %\end{equation}
%%%%%%%%%%%%%%%%%%%%%%%%%%%%%%
%%
%%\text{ where }  
%%%\begin{equation}
%%w^\star_{a,b} =
%%\frac{ \onebb_{a,b \le Q, a\neq b}}{2Q(Q-1)}
%%+ \frac{\onebb_{a\le Q< b} }{2QJ} + \frac{\onebb_{b\le Q<a} }{2QJ} + \frac{ \onebb_{a,b > Q, a\neq b}}{2J(J-1)}.
%WWW
%\begin{cases}
%1/(2Q(Q-1)) & \text{ if } a,b \le Q, a\neq b\\
%1/(2QJ) & \text{ or } a\le Q, b >Q \text{ or } a>Q, b\le Q \\
%1/(2J(J-1)) &  \text{ or } a,b >Q, a\neq b
%\end{cases}.
\frac{1}{2}\begin{pmatrix}
-\frac{\onebb_{Q\times Q}- Id_Q}{Q(Q-1)} & \frac{\onebb_{Q\times J}}{QJ} \\
\frac{\onebb_{J\times Q}}{QJ} & -\frac{\onebb_{J\times J}- Id_J}{J(J-1)}
\end{pmatrix}=:w^\star,
\end{equation}
where the last identity is a notation and we used the usual conventions that for any positive integers $n$, $n_1$, $n_2$  the identity matrix in dimension $n$ is $Id_n$ and $\onebb_{n_1 \times n_2}$ is the matrix with
$n_1$ lines and $n_2$ columns and all entries equal to one. 
Note that if \eqref{eq:blue_sum_over_permutations} is true the conclusion follows because $w^\star$  corresponds precisely to the estimator in \eqref{eq:optimal_blue_estimator_def} and
\eqref{eq:blue_iterm1_variance_ineq} informs that its variance is lower than that of any other unbiased linear estimator.

In order to prove \eqref{eq:blue_sum_over_permutations}, suppose for instance that $a,b \le Q$ (all other situations are analogous). Then~: 
\begin{eqnarray}
	& \  &
\left(\frac{\sum_{\pi \in \Scal_Q, \rho \in \Scal_J} w_{\pi \otimes \rho}}{Q! J!}\right)_{a,b} =  
\left(\frac{J!\sum_{\pi \in \Scal_Q} w_{\pi(a),\pi(b)}}{Q!J!}\right)_{a,b} =  
\sum_{\substack{a_0,b_0 \le Q \\b_0 \neq a_0}}
\sum_{\substack{\pi \in \Scal_Q, \\\pi(a)= a_0, \pi(b)=b_0}}
\left(\frac{w_{a_0,b_0}}{Q!}\right) 
\nonumber \\ & \ & 
\sum_{\substack{a_0,b_0 \le Q \\b_0 \neq a_0}}
%\sum_{\substack{\pi \in \Scal_Q, \\\pi(a)= a_0, \pi(b)=b_0}}
(Q-2)!\frac{w_{a_0,b_0}}{Q!}= -\frac{1}{2} \cdot \frac{1}{Q(Q-1)},
\end{eqnarray}
where for the last identity we used \eqref{eq:constraints_bias_zero_blue}.

\noindent {\bf Item~\ref{item:blue1sample}:} The proof is similar to that of item~\ref{item:blue1sample}.  Note that the terms
$d^2(\delta_{Z_j},\delta_{Z_{j'}})$ appear twice in  \eqref{eq:optimal_blue_estimator_def_1sample}, once with 
 coefficient $-\frac{1}{2J^2}$ and another time 
 with coefficient $-\frac{1}{2J^2(J-1)}$ which sum up to
  $-\frac{1}{2J(J-1)}$ appearing in \eqref{eq:optimal_blue_estimator_def}.
\end{proof}
%%%%%%%%%%%%%%%%%%%%%%%%
\subsection{Proof of the Proposition~\ref{prop:mean_distance_and_decay}} \label{sec:proof_prop:mean_distance_and_decay}

\begin{proof}
The equation \eqref{eq:decay_uniform} is a particular case of \eqref{eq:decay_general} obtained by direct replacement of the values $\alpha_j$; the last point of the conclusion is a direct consequence of \eqref{eq:decay_general} and of the fact that $\alpha\in \Pcal_J$ i.e., all are positive and sum up to one. Thus, all that remains to be proved is \eqref{eq:decay_general}.
We recall the formula (see~\cite{szekely_energy_2013,rkhs_book_berlinet_agnan,turinici_radonsobolev_2021}, compare also with formula~\eqref{eq:defdistance})~:
\begin{equation}
    d(\nu,\xi)^2 =\Ebb_{X\sim\nu, Y\sim\xi, X\independent Y}h(X,Y) - 
    \frac{\Ebb_{X,X'\sim\nu, X\independent X'}h(X,X')}{2}-
    \frac{\Ebb_{Y,Y'\sim\nu, Y\independent Y'}h(Y,Y')}{2}.
\end{equation}
In particular the other hand for $Z\sim\mu$~:
\begin{eqnarray}
& \ & 
\Ebb_{Z \sim \mu} \left[ d\left(\delta_Z ,\mu\right)^2  \right]= 
\Ebb_{Z \sim \mu} \left[ \Ebb_{X\sim\mu,  X\independent Z}h(X,Z)
-  \frac{\Ebb h(Z,Z)}{2}
-  \frac{\Ebb_{Y,Y'\sim\mu, Y\independent Y'}h(Y,Y')}{2}\right].
\nonumber \\ & \ & 
=\frac{\Ebb_{Y,Y'\sim\mu, Y\independent Y'}h(Y,Y')}{2} =: \frac{v}{2},
\end{eqnarray}
where the last equality is a notation.
Since $h(x,y) = d^2(\delta_x,\delta_y)$, by renoting $Y,Y'$ as $X,Y$ we can write~:
\begin{equation}
    \Ebb_{X,Y \sim \mu, X \independent Y} \left[ d\left(\delta_X ,\delta_Y\right)^2  \right]= 
    \Ebb_{X,Y \sim \mu, X \independent Y} \left[ h(X,Y) \right]=     v.
\end{equation}

Making use of this relation and of the ``parallelogram" identity \eqref{eq:parallelogram} 
we write~:
\begin{eqnarray}
& \ & 
\Ebb \left[ d\left(\delta_{\alpha,X} ,\mu\right)^2  \right]
= \Ebb \left[ \sum_{j=1}^J \alpha_j d\left(\delta_{X_j} ,\mu\right)^2
- \frac{1}{2}\sum_{j,j'=1}^J \alpha_j \alpha_{j'} d\left(\delta_{X_j} ,\delta_{X_{j'}} \right)^2
\right]
\nonumber \\ & \ &
= \frac{v}{2} \sum_{j=1}^J \alpha_j  - \frac{v}{2} \sum_{j,j'=1, j \neq j'}^J \alpha_j \alpha_{j'} 
\stackrel{\alpha \in \Pcal_J}{=\joinrel=\joinrel=}
\frac{v}{2}
\left( 1-  \sum_{j,j'=1}^J \alpha_j \alpha_{j'}  +
 \sum_{j=1}^J \alpha_j^2 
\right)
\nonumber \\ & \ & =
\frac{v}{2}
\left( 1-  \left(\sum_{j=1}^J \alpha_j \right)^2 +
\sum_{j=1}^J \alpha_j^2 
\right)
=\frac{v}{2}\left(1-1 +
\sum_{j=1}^J \alpha_j^2 
\right)
=\frac{v}{2}\sum_{j=1}^J \alpha_j^2, 
\end{eqnarray}
which ends the proof.
\end{proof}

%%%%%%%%%%%%%%%%%%%%%%%%%%%%%%%%%%%%%%%%%%%
\subsection{Proof of Proposition~\ref{prop:uniqueweights}} \label{sec:proof_prop:uniqueweights}

\begin{proof}
Let $\alpha_a$, $a=1,2$ correspond to two optimal quantizations of $\mu$ with same points $X$ and weights $\alpha_a$. Since they are optimal they will have same distance to $\mu$ which is minimal that we denote $d_{min}$: $d(\mu,\delta_{\alpha_a,X})=d_{min}$, $a=1,2$.
But, from the parallelogram identity:
\begin{eqnarray}& \ & 
d(\mu,\delta_{t\alpha_1+(1-t)\alpha_2,X})^2=
d(\mu,t\delta_{\alpha_1,X} + (1-t)\delta_{\alpha_2,X})^2 
 \nonumber \\ & \ &
= t d(\mu,\delta_{\alpha_1,X})^2  +
(1-t) d(\mu,\delta_{\alpha_1,X})^2-t(1-t) d(\delta_{\alpha_1,X},\delta_{\alpha_2,X})^2 \le d_{min}, 
\end{eqnarray}
with equality only when $\delta_{\alpha_1,X}= \delta_{\alpha_2,X}$.
\end{proof}
%%%%%%%%%%%%%%%%%%%

\subsection{Proof of Proposition~\ref{prop:1D_quantiles}} \label{sec:proof_prop:1D_quantiles}
\begin{proof}
Under our hypothesis the cumulative distribution function $F_\mu :\R \to ]0,1[$ of $\mu$ is invertible and strictly increasing;
we denote by $q^\mu_r$ the quantile of order $r$ of the law $\mu$, i.e. $q^\mu_r = F_\mu^{-1}(r)$.

Let $X \in \R^J$ be an optimal solution of the minimization problem. Denote $y_j$ the $j$-th value in $X$ after ordering, i.e. 
such that there are $j-1$ values in $X_j$ below $y_j$ and 
$J-j$ above it (except if some other point $X_l$ is equal to $y_j$, for now we suppose this is not the case). Note that moving $y_j$ to $y_j+\Delta X$ (the others remain identical) will produce 
a new point $X'$
and, when $\Delta X>0$, the following change in the distance
\begin{equation}
 d(\delta_{\alpha,X},\mu)^2 -    
 d(\delta_{\alpha,X'},\mu)^2  = \frac{2 \Delta X}{J} \left( \left[1- F_\mu(y_j) - \frac{J-j}{J}\right] -  \left[F_\mu(y_j)- \frac{j-1}{J} \right]
 \right) + o(\Delta X).
\end{equation}
Because of the optimality, this quantity cannot be negative, which shows that $F(y_j) \ge \frac{j+1/2}{J}$.
A similar analysis shows that 
 $F(y_j) \le \frac{j+1/2}{J}$ so 
 finally $F(y_j) = \frac{j-1/2}{J}$, which is our conclusion.
\end{proof}
 
%%%%%%%%%%%%%%%%%%%%%%%
\subsection{Remarks on the applications of the Proposition ~\ref{prop:existence_fixed_alpha}}
\begin{remark}
In practice, when $\eta$ is a positive measure
 it is natural to choose $\alpha \in (\R_+)^Q$ such that 
$\sum_q \alpha_q = \int \eta(dx)$; the situation is, up to some multiplicative constants, identical to the quantization of a probability law; the minimal distance will be zero when $\eta$ is a positive sum of Dirac measures and $Q$ is large enough. This situation is covered by the Proposition~\ref{prop:existence_fixed_alpha}.
On the contrary, when $\eta$ is a signed measure two main use cases can appear~:

- when $\eta$ is a positive measure  
and $\delta_{\beta,Y}$ represents a previous attempt at measure quantization,
such that 
$\int \eta_+(dx) \ge \int \delta_{\beta,Y}(dx)$ we only seek to refine  this already available ``historical" fixed part. Then it is natural to choose  again $\alpha \in (\R_+)^Q$ and
$\sum_q \alpha_q = \int \eta(dx)- \lambda  \int \delta_{\beta,Y}(dx)$ 
with $\lambda \in[0,1]$
and quantize the 
(probably non-positive, i.e., signed) measure $\eta_+(dx) - \lambda \delta_{\beta,Y}$; the parameter $\lambda$ is chosen by the user.
This situation is also covered by the Proposition~\ref{prop:existence_fixed_alpha}.
%and illustrated numerically in Section~\ref{sec:mnist_increment_signed}.

- when $\eta$ is intrinsically a non positive signed measure, 
 with the canonical (Jordan) decomposition as difference of 
two positive measures $\eta=\eta_+ - \eta_-$,
then we may want to quantize it the best we can, and in this case $\alpha$ will be chosen with possibly negative parts and such that
$\sum_q |\alpha_q| = \int |\eta|(dx)= \int (\eta_++\eta_-)(dx)$.
This situation is not covered by the Proposition~\ref{prop:existence_fixed_alpha}.
Of course, one possibility is to quantize $\eta_+$ and $\eta_-$ separately with positive weights if the decomposition is known (or can be sampled).
\end{remark}
%\begin{remark} The hypotheses \eqref{eq:hyp_lowerbound_h}-\eqref{eq:coercivity} are satisfied in practice by many kernels  and in particular by the kernels we advocate $\sqrt{a^2+|x|^2}- a$ for any $a\ge 0$. Indeed, the coercivity is obvious, and the lower bound is a consequence of immediate inequalities with $C_p=0$. Note that, on the contrary, the kernel $k$ is not upper bounded. In addition, the hypothesis \eqref{eq:lowerbound} may be weakened but at the price of additional technicalities.\end{remark}

\subsection{Remarks on the existence for negative weights}

We consider here the situation when the weights $\alpha$ can be chosen negative. 
When the kernel $k$ is bounded and the domain $\Xcal$ is also bounded, standard techniques allow to prove the existence of the optimal quantizer.
However,  when the kernel $k$ is unbounded or the domain $\Xcal$ is unbounded the question of the existence of the optimal quantization needs careful consideration as illustrated in the example below.

\begin{example} Consider  the energy kernel
and 
$\xi_n=n\delta_{\sqrt{n}+1/n^3}-n\delta_{\sqrt{n}} + \delta_a$; note that  $d(\xi_n,\delta_0)^2 \to d(\delta_a,\delta_0)^2$; in particular when $a=0$ this distance goes to zero, but note that the total variation norm explodes and the support of the measure is not bounded. Any such sequence can be added to any minimizer sequence to perturb its total variation norm and support without perturbing its minimizing character. 
\label{ex:distance_zero_but_support_not_close}
\end{example}

\subsection{Remarks concerning the moments}

As we saw in Proposition~\ref{prop:1D_quantiles}
 the quantization can be related to quantile information concerning the target measure, which is generally understood as a ``zero-th" order moment. It does not, in general, ensure precise reconstruction of moments of higher order, in particular of the mean; for instance for $Q=1$ and the `energy' kernel (i.e., $\dfrak_{1,0}^{HE}$) the quantization point will be the median and not the mean (the mean will be recovered in the limit $Q\to\infty$).
When the mean is important one can improve the quantization properties; two cases appear~:

- either the mean is known (e.g., for applications in physics where the total energy is known and has to be conserved exactly): in this situation one can look for minimizers only among those matching the correct mean (same for any other relevant statistics such as the variance)

- the mean is unknown~: in this case 
a good value of the mean can be enforced by replacing  the kernel $h_{r,a}^{HE}$ by the kernel 
$h_{r,a}^{HE} + \lambda \cdot h_{2,0}^{HE}$; when the penalization parameter $\lambda >0$ is large enough the minimizer will tend to  have a small error in the 
metric  $\dfrak_{2,0}^{HE}$; recall that  
$\dfrak_{2,0}^{HE}(\mu,\nu) = (\Ebb{\mu}- \Ebb{\nu})^2$
therefore such a kernel will contribute towards reproducing the mean of the target distribution.
Similar considerations hold for any statistical objects depending on the measure.

%%%%%%%%%%%%%%%%ù

\subsection{Positivity of the quantization}

A relevant question is also whether the quantization of a positive measure will also remain positive. We do not have a complete general answer at the moment but it is obvious that such a result would require some hypothesis as illustrated by the example below. At the very least the optimal points should be allowed to move freely in the convex hull of the support of the measure.

\begin{figure}[!htb]
    \centering
    \includegraphics[width=0.3\textwidth]{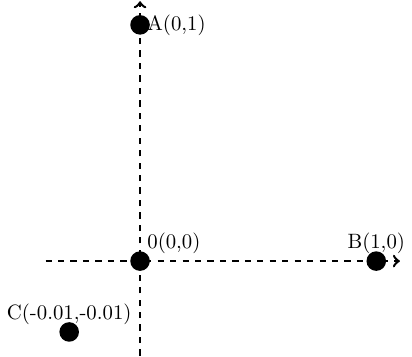}
%    \begin{tikzpicture}[x=4cm,y=4cm]
%    \draw[->,very thick,dashed] (-0.4,0)--(1.1,0); % l'axe des abscisses
%    \draw[->,very thick,dashed] (0,-0.4)--(0,1.1); % l'axe des ordonnées
%    \draw (-0.3,-0.3) node[anchor=south] {C(-0.01,-0.01)};
%    \draw (0,0) node[anchor=south west] {0(0,0)};
%    \draw (0,1) node[anchor=west] {A(0,1)};
%    \draw (1,0) node[anchor=south] {B(1,0)};
%    \foreach \Point in {(-0.3,-0.3), (0,0), (0,1), (1,0)}{
%        \node at \Point {\textbullet};
%    }
%  \end{tikzpicture}
    \caption{Illustration of the support of the target measure $\mu$ in Example~\ref{ex:non_positive_weights}. For visual reasons the axis scales are not uniform (otherwise the point $C$ would be difficult to distinguish from $O$).}
    \label{fig:counter_ex_positive_alpha}
\end{figure}

\begin{example}[non positivity of the quantization]
Consider a $N=2$ dimensional target measure $\mu=
\left(\frac{1}{3}-\epsilon\right)\delta_0+ 
\left(\frac{1}{3}-\epsilon\right)\delta_A+ 
\left(\frac{1}{3}-\epsilon\right)\delta_B+ 
3\epsilon\delta_C$  consisting of Dirac masses
at points $O(0,0)$, $A(0,1)$, $B(1,0)$ and $C(-0.01,-0.01)$ (the coordinates are given in parenthesis, see Figure~\ref{fig:counter_ex_positive_alpha} for an illustration). We set $\epsilon=0.001$.
Suppose we want to quantize this measure with $Q=2$ points and we also have the restriction to only look for points in the {\bf support} of the target measure
i.e. only consider points $0$, $A$, $B$ or $C$. The distance is $\dfrak_{1.95}$.
Therefore we look for the optimum of 
\begin{equation}
\alpha\in \R, X,Y \in \{0,A,B,C\} \mapsto 
 \dfrak_{1.95} \left(\mu, \alpha \delta_X + (1-\alpha) \delta_Y \right)^2.
\end{equation}
One can compute the optimal quantization by enumerating all pairs of admissible points $X,Y$ and optimizing the weight parameter $\alpha$ (the problem is a 1D quadratic optimization). We find that the minimal distance is realized by 
$X^\star=0$, $Y^\star=C$ and the optimal weight is $\alpha^\star=27.108$
and $1-\alpha^\star=-26.108$ which is negative.
\label{ex:non_positive_weights}
\end{example}

\end{document}